\pdfoutput=1
\documentclass[11pt]{article}

\usepackage{amsmath}
\usepackage{microtype}
\usepackage{xcolor}
\usepackage{adjustbox}
\usepackage{graphicx}
\usepackage{subfigure}

\usepackage{booktabs} 
\usepackage[skip=0pt]{caption} 
\usepackage[hyphens]{url}
\usepackage{hyperref}
\usepackage[utf8]{inputenc}
\usepackage{array}
\usepackage{float}

\usepackage{listings}
\usepackage{titlesec}

\lstdefinelanguage{Prompt}{
    basicstyle=\ttfamily\small,
    frame=single,
    backgroundcolor=\color{gray!10},
    breaklines=true,              
    keywordstyle=\color{blue}\bfseries,    
    stringstyle=\color{red},             
    commentstyle=\color{green!50!black},  
    morekeywords={CREATE, TABLE, SELECT, WHERE, AS, CASE, WHEN, THEN, ELSE, END, FROM, ORDER, BY, LIMIT},
    literate={\{}{{\textcolor{cyan}{\{}}}{1} 
             {\}}{{\textcolor{cyan}{\}}}}{1} 
             {self.table}{{\textcolor{magenta}{self.table}}}{10} 
             {self.question}{{\textcolor{magenta}{self.question}}}{13} 
             {self.description}{{\textcolor{magenta}{self.description}}}{16}
             {self.table.columns}{{\textcolor{magenta}{self.table.columns}}}{18} 
             {df.column}{{\textcolor{magenta}{df.column}}}{9} 
             {self.plan}{{\textcolor{magenta}{self.plan}}}{9} 
             {self.plan}{{\textcolor{magenta}{self.plan}}}{9} 
             {step.prompt}{{\textcolor{magenta}{step.prompt}}}{11} 
             {self.name}{{\textcolor{magenta}{self.name}}}{9}
             {Example}{{\textbf{Example}}}{7}
             {MySQL_Code_Generation:}{{\textbf{MySQL Code Generation:}}}{22} 
             {Instructions:}{{\textbf{Instructions:}}}{12} 
             {Step_1}{{\textbf{Step 1}}}{6}
             {Step_2}{{\textbf{Step 2}}}{6}
             {Step_3}{{\textbf{Step 3}}}{6}
             {Step_4}{{\textbf{Step 4}}}{6}
             {Step_5}{{\textbf{Step 5}}}{6}
             {Step_6}{{\textbf{Step 6}}}{6}
             {Question_:}{{\textbf{Question:}}}{9}
             {LLM_Step}{{\textbf{LLM Step}}}{8}
             {SQL_Step}{{\textbf{SQL Step}}}{8}
             {Champion_}{{\textcolor{red}{'\%Champion\%'}}}{13} 
        {Win__}{{\textcolor{red}{'Win'}}}{5}
        {Round__}{{\textcolor{red}{'\%1st Round\%'}}}{14}
        {No_Win}{{\textcolor{red}{'No Win'}}}{8}
        {1936_}{{\textcolor{red}{'1936'}}}{6}
}

\lstdefinelanguage{SQL}{
  morekeywords={CREATE, TABLE, SELECT, CASE, WHEN, THEN, ELSE, END, IS, NULL, LIKE, SUBSTRING, CAST, AS, REGEXP_REPLACE},
  sensitive=true,
  keywordstyle=\color{blue}\bfseries,
  commentstyle=\color{green!60!black},
  stringstyle=\color{red},
  basicstyle=\ttfamily\footnotesize,
  moredelim=[is][\color{purple}]{`}{`}, 
  literate={CREATE}{{\color{blue}\textbf{CREATE}}}{5}
           {TABLE}{{\color{blue}\textbf{TABLE}}}{5}
           {SELECT}{{\color{blue}\textbf{SELECT}}}{6}
           {CASE}{{\color{blue}\textbf{CASE}}}{4}
           {WHEN}{{\color{blue}\textbf{WHEN}}}{4}
           {THEN}{{\color{blue}\textbf{THEN}}}{4}
           {ELSE}{{\color{blue}\textbf{ELSE}}}{4}
           {END}{{\color{blue}\textbf{END}}}{3}
           {IS}{{\color{blue}\textbf{IS}}}{2}
           {NULL}{{\color{purple}\textbf{NULL}}}{4}
           {LIKE}{{\color{purple}\textbf{LIKE}}}{4}
           {SUBSTRING}{{\color{purple}\textbf{SUBSTRING}}}{9}
           {CAST}{{\color{purple}\textbf{CAST}}}{4}
           {REGEXP_REPLACE}{{\color{purple}\textbf{REGEXP\_REPLACE}}}{13}
}

\usepackage[final]{ACL2023}

\usepackage{times}
\usepackage{latexsym}

\usepackage[T1]{fontenc}

\usepackage{inconsolata}

\title{Weaver: Interweaving SQL and LLM for Table Reasoning}

\newcommand{\methodName}{{Weaver~}}

\author{Rohit Khoja$^{1}$$^\textbf{*}$ \quad Devanshu Gupta$^{1}$$^\textbf{*}$ \quad \textbf{Yanjie Fu}$^{1}$ \quad \textbf{Dan Roth}$^{2}$ \quad \textbf{Vivek Gupta}$^{1}$\textsuperscript{\textdagger} \\
        $^{1}$Arizona State University $^{2}$University of Pennsylvania \\
        \texttt{\{rkhoja2,dgupta77,yanjiefu,vgupt140\}}@asu.edu \quad \texttt{danroth}@seas.upenn.edu
        }

\begin{document}
\maketitle
\begingroup\def\thefootnote{}\footnotetext{$\textbf{*}$These authors contributed equally to this work. \quad \quad \textdagger primary supervisor of this work.}\endgroup
\begin{abstract}

Querying tables with unstructured data is challenging due to the presence of text (or image), either embedded in the table or in external paragraphs, which traditional SQL struggles to process, especially for tasks requiring semantic reasoning. While Large Language Models (LLMs) excel at understanding context, they face limitations with long input sequences. Existing approaches that combine SQL and LLM typically rely on rigid, predefined workflows, limiting their adaptability to complex queries. To address these issues, we introduce \methodName, a modular pipeline that dynamically integrates SQL and LLM for table-based question answering (Table QA). \methodName generates a flexible, step-by-step plan that combines SQL for structured data retrieval with LLMs for semantic processing. By decomposing complex queries into manageable subtasks, \methodName improves accuracy and generalization. Our experiments show that \methodName consistently outperforms state-of-the-art methods across four Table QA datasets, reducing both API calls and error rates.

\end{abstract}

\section{Introduction}
Tables play a critical role across various domains such as finance (e.g., transaction records), healthcare (e.g., medical reports), and scientific research. However, many real-world tables often contain a mix of structured fields alongside columns with embedded unstructured text (such as free-form text or images), which makes reasoning and information retrieval challenging. Extracting insights from such data demands both logical and semantic reasoning. While SQL and Python-based methods excel in handling structured data, they fall short in dealing with unstructured text, missing entries, or implicit inter-column relationships.

\begin{figure}[H]
    \centering
  \includegraphics[width=0.48\textwidth]{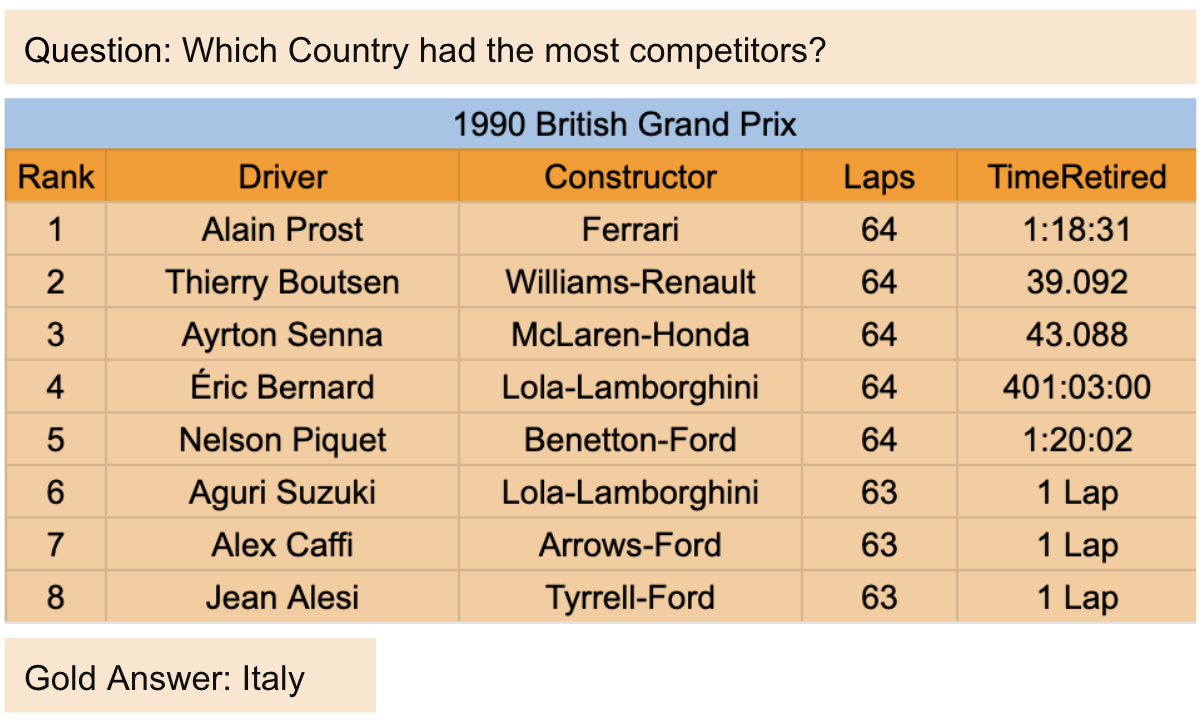}
    \caption{\small
    An exemplar hybrid query from the WikiTQ Dataset demonstrating the need for interwoven reasoning. The question “Which country had the most competitors?” requires the semantic inference of driver nationalities and the logical grouping and counting of competitors to produce the correct answer.}
    \label{fig:1}
\vspace{-0.75em}
\end{figure}

Recent advances in Large Language Models (LLMs) have demonstrated strong capabilities in natural language understanding and contextual reasoning, opening new avenues for complex tasks. However, LLM still face key limitations, particularly with long contexts and numerical or temporal reasoning. For instance, in the WikiTableQuestions \cite{wikitq} dataset, the query “Which country had the most competitors?” Figure \ref{fig:1} requires inferring the competitors' countries from a \texttt{driver} column information not explicitly present. While traditional tools such as SQL or Python cannot resolve such gaps, LLM can leverage their pre-trained knowledge to do so. Yet, the subsequent grouping and counting by country is something LLM struggle with but SQL handles well. Solving such queries effectively requires a hybrid approach that combines the strengths of LLM and programmatic methods. This raises a key question: \textit{Can SQL and LLM be seamlessly interwoven?}

Some methods, such as Binder \cite{binder} and BlendSQL \cite{BlendSQL}, integrate LLM into SQL workflows by treating them as function calls. For example, Binder combines LLM reasoning with SQLite to support hybrid queries. While effective for simpler tasks, these approaches struggle with complex queries, as LLM often fail to generate accurate multi-step logic. This highlights the need to decompose complex queries into smaller, manageable steps. Other approaches, such as H-STAR \cite{hstar} and ReAcTable \cite{react}, use programmatic techniques to prune tables but rely heavily on costly API calls. Meanwhile, methods like ProTrix \cite{protrix} limit reasoning to just two steps, making them insufficient for multi-hop queries. These rigid pipelines often constrain LLM to answer extraction and cannot handle questions requiring external or implicit knowledge.

To address these challenges, we introduce a modular, planning-based framework that dynamically alternates between SQL for logical operations and LLM for semantic reasoning. By decoupling these components, our approach overcomes the limitations of monolithic systems and significantly improves performance on complex Table QA tasks. The process begins with extracting relevant columns and generating column descriptions based on the given table and query, resolving formatting inconsistencies and ambiguities in table or column names. An LLM then generates a step-by-step reasoning plan, combining SQL queries for structured operations with LLM prompts for semantic inference or column augmentation. Each step produces an intermediate table, enabling transparent reasoning and easy backtracking. A final answer extraction step retrieves the result from the processed table. This adaptable design enables seamless integration with various database engines (e.g., MySQL, SQLite). Our approach offers the following key contributions:

\begin{itemize}
\setlength\itemsep{0.0em}

    \item We propose \textit{\methodName}, a modular and interpretable framework for hybrid query execution that dynamically decomposes complex queries into modality-specific steps (e.g., SQL, LLM, vision-language models (VLMs)) without manual effort.

    \item We conduct extensive experiments on multiple hybrid QA benchmarks, including multimodal datasets, showing that \textit{\methodName} outperforms existing methods by large margins, particularly on complex, multi-hop reasoning queries.

    \item We introduce a query plan optimization strategy that improves execution efficiency with minimal accuracy loss. \textit{\methodName} also stores all intermediate outputs, enabling transparency, effective human-in-the-loop debugging.
\end{itemize}

\vspace{-0.5em}
Our results show superior accuracy over existing baselines, especially for queries requiring implicit reasoning beyond explicit table values. By bridging structured and unstructured reasoning, our approach sets a new benchmark for complex Table QA, offering a scalable solution for real-world applications.

The code, along with other associated scripts, are available at \url{https://coral-lab-asu.github.io/weaver}.

\begin{figure*}[h]
\centering
\includegraphics[width=0.99 \textwidth]{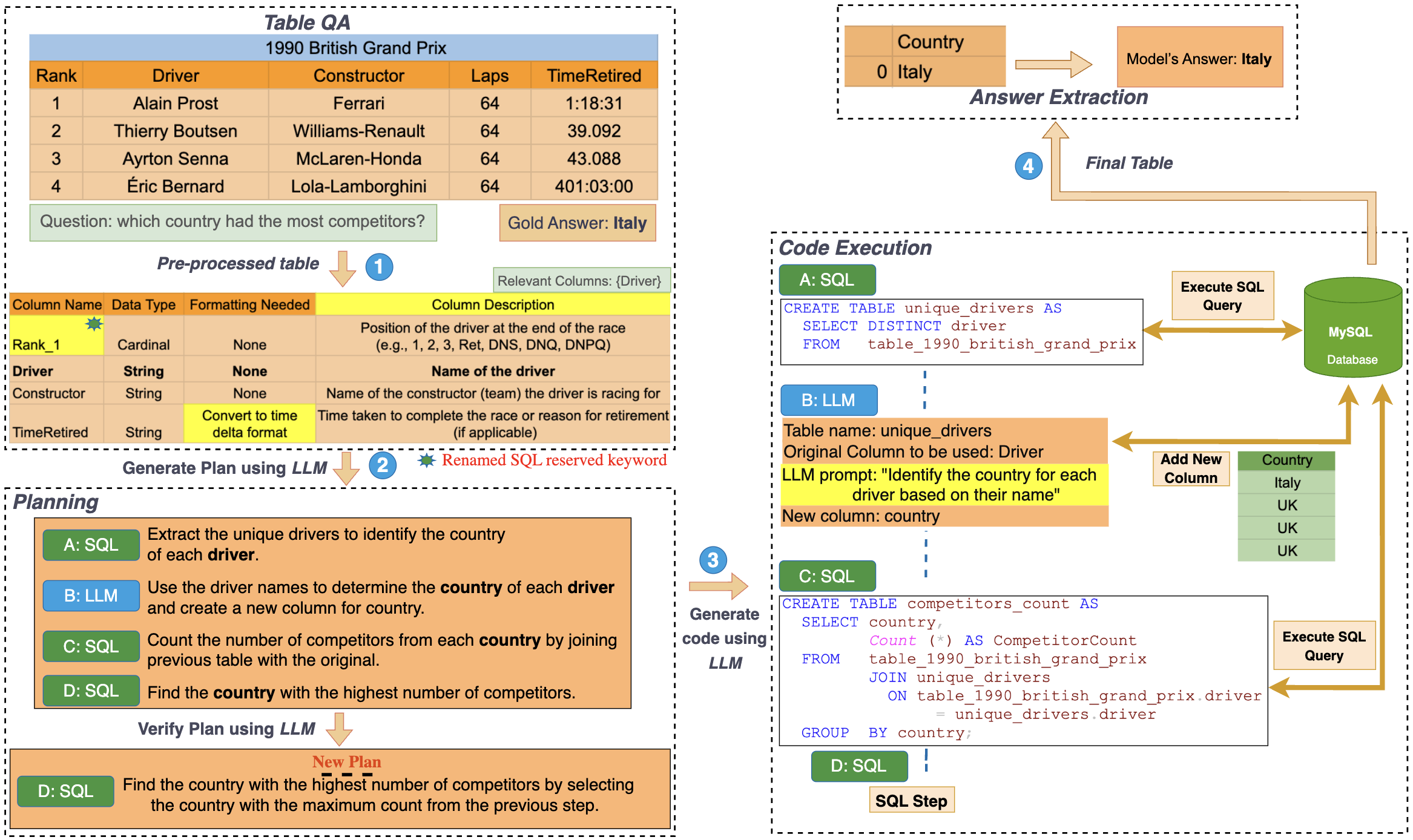}
\caption{\small The transparent and interpretable \methodName pipeline for Table QA. The framework dynamically alternates between SQL-based structured data operations (e.g., creating the \texttt{unique\_drivers} table) and LLM-based semantic reasoning (e.g., inferring  ‘country’ from \texttt{driver} names), with each step producing a traceable intermediate table to mitigate errors and enhance debugging.}
\label{fig:main_figure}
\vspace{-0.75em}
\end{figure*}

\section{Our Approach}
\label{section:our-approach}

    This study addresses question answering tasks over tables containing both structured and unstructured data. Each instance consists of a table ($T$), a user query ($Q$), an optional paragraph ($P$), and a predicted answer ($A$). Queries span multiple categories: short-form queries (WikiTQ) involving direct lookups or aggregations; fact-checking queries (TabFact \cite{tabfact}) that require claim verification; numerical reasoning queries (FinQA \cite{finqa}) necessitating multi-step calculations; and multimodal queries (FinQA and OTT-QA \cite{ott-qa}) that demand reasoning over extensive textual contexts inside and outside tables. Complex queries, such as “Which country had the most competitors?”, frequently require semantic inference when explicit data is unavailable. Previous approaches typically rely on single-step executions, limiting flexibility and interpretability. In contrast, our method dynamically integrates SQL for structured data operations and LLM for semantic inference, providing adaptable and accurate query resolution.

\subsection{SQL-LLM \methodName}

We introduce \methodName, a novel methodology integrating SQL and LLM specifically designed for Table QA tasks involving complex semantic reasoning and free-form responses. \methodName operates through distinct, structured phases:

\paragraph{1. Pre-processing:}  
We begin by preprocessing the tables to mitigate SQL-related errors due to naming conflicts and data inconsistencies. This involves renaming columns conflicting with SQL reserved words (e.g., \texttt{Rank}), removing special characters, and standardizing column names. Subsequently, an LLM identifies and extracts relevant columns for the query, generating descriptive metadata for these columns. This metadata clarifies schema interpretations, resolves formatting issues, and defines accurate data types, as illustrated in Figure \ref{fig:main_figure}. For external unstructured text, relevant information is retained using an LLM to ensure context alignment.

\paragraph{2. Planning:}  

In this phase, an LLM generates a dynamic, step-by-step plan using few-shot prompting based on the previously derived metadata. The plan consists of sequential subtasks, each categorized explicitly as either SQL or LLM operations. 

\paragraph{(a.) {SQL Step}:}
SQL steps manage structured data tasks, including filtering rows, formatting column data types, mathematical operations and data aggregations. For example, Figure \ref{fig:main_figure} demonstrates an SQL step that generates an intermediate table, \texttt{unique\_drivers}.

\paragraph{(b.) {LLM Step}:}  
LLM steps handle tasks beyond SQL's capabilities, such as deriving new columns through semantic inference, sentiment analysis, or interpreting complex textual data. LLM leverage either contextual paragraphs or their pretrained knowledge to perform these tasks. Each LLM step carefully integrates outputs back into the structured data tables to ensure coherence and consistency for subsequent SQL steps. Figure \ref{fig:main_figure} illustrates how an LLM infers a \texttt{country} column from the \texttt{driver} column in the intermediate table \texttt{unique\_drivers}. The LLM is guided through structured prompts that leverage its pretrained knowledge and reasoning capabilities. Relevant information (extracted from external unstructured text in pre-processing step) is also passed to LLM if present.

\paragraph{Plan Verification:}
Prompting techniques like self-refinement \cite{selfrefineiterative} and verification \cite{self-verification} are known to enhance LLM reasoning by reducing errors and improving consistency. To leverage this, we use a secondary LLM to verify the initial plan, ensuring its logical consistency, robustness, and completeness. Gaps such as insufficient reasoning or formatting issues are addressed by refining the plan, as shown in planning step (D) (New Plan) in Figure \ref{fig:main_figure}. This verification improves the pipeline’s reliability and mitigates cascading errors.

\paragraph{3. Code Execution:} Following verification, the pipeline executes the plan sequentially, combining SQL queries and LLM-generated prompts.

\paragraph{(a.) {SQL Step - Query Generation}:} SQL operations involve formatting, filtering, joining, aggregating, and grouping data, with intermediate tables stored at each stage. SQL efficiently handles structured data operations, reducing reliance on LLM steps.

\paragraph{(b.) {LLM Step - Prompt Generation}:} At each LLM step execution, we retrieve the inputs specified in the plan, namely the relevant column subset, source table, LLM prompt, and target column name. Using the prompt and selected columns, the LLM generates the target column values, which are then appended to the intermediate table.

Robust error handling and fallback mechanisms ensure pipeline robustness, utilizing the recent successful intermediate table if execution errors occur.

\paragraph{4. Answer Extraction:}  
In the final pipeline stage, the intermediate table and user query are inputted to an LLM, which generates a natural language answer. Leveraging few-shot learning ensures output consistency and contextual accuracy, effectively resolving complex queries.

Figure \ref{fig:main_figure} illustrates this process with a sample Table QA example.

\subsection{Optimization} We experimented to enhance the efficiency of our pipeline, optimizing the planning strategy by minimizing unnecessary LLM calls and prioritizing SQL-based operations.
\begin{figure}[H]
\vspace{-0.75em}
    \centering
  \includegraphics[width=0.48\textwidth]{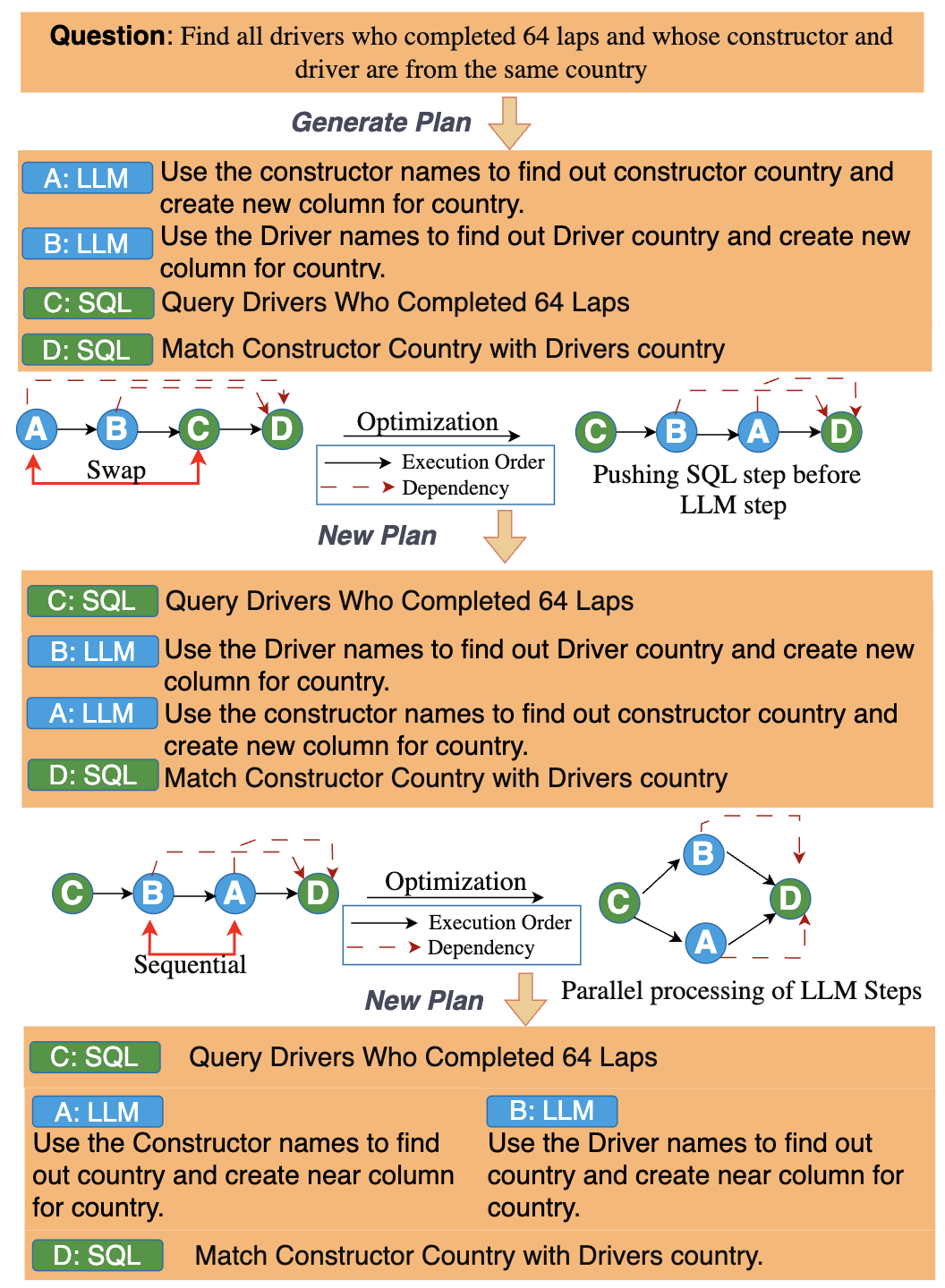}
    \caption{\small Optimization using SQL Reordering and LLM call parallelization.}
    \label{fig:llm-optimizaton-swap}
\end{figure}
One key optimization involves pushing SQL operations such as filtering, aggregation, and formatting early in the pipeline as shown in Figure \ref{fig:llm-optimizaton-swap}. This reduces the volume of data that needs to be processed by the LLM, significantly reducing latency and computational overhead. Furthermore, parallelizing LLM inference across these optimized chunks further enhances efficiency, allowing the system to handle large-scale tabular reasoning tasks effectively, as shown in Figure \ref{fig:llm-optimizaton-swap}. To further streamline execution, sequential SQL steps are merged, reducing SQL calls and improving performance, see Figure \ref{fig:3}.\\
Given the computational demands of LLMs for large tables, we split data into smaller context-aware chunks before sending them to the LLM for inference. This prevents input truncation, maintains logical coherence between batches, and ensures optimal utilization of the context window of the model, as demonstrated in Figure \ref{fig:llm-optimizaton-rows-batch}, in the Appendix \ref{appendix:optmization}. This optimization strategy further enhances \textit{\methodName} planning and execution efficiency.

\section{Experiments}

\paragraph{Benchmarks:}
As hybrid multi-hop Table QA remains an emerging research area, there is no dedicated benchmark to evaluate such tasks. To fill this gap, we curate a hybrid subset by filtering relevant examples from several established table-based datasets for the evaluation of \textit{\methodName}.

\paragraph{{Source Datasets:}}
For a comprehensive evaluation of \textit{\methodName}'s ability to handle complex hybrid queries, we assess its performance across three diverse datasets: WikiTQ (a short-form answering dataset), TabFact (a fact verification dataset), and FinQA (a numerical reasoning dataset). We also evaluated our approach on 3,000 queries each of multimodal (MM) datasets, FinQA$_{\text{MM}}$ and OTT-QA$_{\text{MM}}$, which require reasoning in both tabular data, and the accompanying textual context (usually a paragraph outside tables). We also evaluated \methodName on the MMTabQA$_{\text{MM}}$ dataset \cite{mmtab,titiya2025mmtbench}, which involves reasoning over tables that include both text and images. This dataset contains 1,600 queries and 206 tables, with each query requiring the integration of textual and visual reasoning, including SQL, LLM, and VLM calls. Unlike traditional table QA tasks, these benchmarks challenge the model to integrate information from structured (tables) and unstructured (text, images) modalities. Details on the datasets in Appendix \ref{appendix:dataet_details}.

\paragraph{{Filtering Methodology:}}
We define \textit{hybrid} queries as those that require both SQL operations and LLM-based reasoning. These queries are more complex, as they necessitate not only structured data retrieval but also advanced reasoning capabilities, such as entity inference or free-text interpretation, which SQL alone cannot provide. To identify such queries, we use Binder-generated queries that incorporate \textit{user-defined LLM functions (UDFs)}. Queries involving UDFs indicate the need for semantic reasoning beyond SQL’s capabilities. We did not validate the correctness of Binder's outputs; its role was purely to flag queries with hybrid characteristics. For $\text{FinQA}$, we utilized the ‘qa’ object to identify queries requiring multiple reasoning steps, excluding simple table lookups to ensure the selected queries involved more than just direct data retrieval. All candidate queries were then manually validated to confirm they require both SQL execution and LLM reasoning steps.

\paragraph{{Dataset Statistics:}}
After filtering, the hybrid versions of the datasets consist of: (a) \textsc{WikiTQ}: 510 examples (original: 4,344), (b) \textsc{TabFact}: 303 examples (original: 2,000), and (c) \textsc{FinQA}: 1,006 examples (original: 8,281). These represent the final queries filtered to create the hybrid versions: $\text{WikiTQ}_{\text{hybrid}}$, $\text{TabFact}_{\text{hybrid}}$, and $\text{FinQA}_{\text{hybrid}}$.

\paragraph{Evaluation Metrics:}
Traditional Exact Match (EM) requires the model’s output to exactly match the gold answer, which can unfairly penalize semantically correct responses that differ only in formatting, such as \textit{2nd April 2024} versus \textit{04/02/2024}. To address this limitation, we introduce the Relaxed Exact Match (REM) metric, which applies a three-step evaluation framework. First, we automatically standardize the model’s output format to align with the gold answer, handling differences such as units, date formats, casing, and common abbreviations. This normalization is performed using an LLM-based transformation prompt (see  Appendix \ref{appendix:prompt_and_examples}), ensuring consistent resolution of trivial mismatches, for example, if the gold answer is \textit{17 years} and the model outputs \textit{17}, the unit is automatically appended to match the reference. Once both answers are format-aligned, we apply the standard EM metric to the normalized outputs, preserving the objectivity and reproducibility of exact string matching. Finally, in the rare cases where discrepancies remain, we incorporate a human safeguard to verify correctness and ensure that automatic normalization has not introduced errors or missed subtle equivalences. Importantly, this human evaluation is not a subjective judgment, but a strict check of semantic equivalence; for example, if the model outputs \textit{5 feet 10 inches} and the gold answer is \textit{70 inches}, the evaluator confirms whether the two represent the same quantity. By combining automatic normalization, objective exact matching, and a minimal safeguard, REM provides a fairer and more reliable evaluation compared to traditional EM.

\paragraph{LLM Models:}
In our research, we use state-of-the-art large language models (LLMs) such as Gemini-2.0-Flash \cite{gemini2}, GPT-4o-mini-2024-07-18, GPT-4o-2024-08-06 \cite{gpt4ocard} and the open-source DeepSeek-R1-Distill-Llama-70B \footnote{\url{https://github.com/meta-llama/llama-models/blob/main/models/llama3\_3/LICENSE}} \cite{deepseek}, \cite{llama} for table reasoning tasks. Our model inputs include in-context examples, the table, and the question for each step of the pipeline. We use the same LLM across all stages of the pipeline (planning, plan verification, code execution and final answer extraction), though the framework can be easily adapted to assign different LLMs to specific tasks. To ensure deterministic and stable output, we use a fixed temperature setting of 0.01.

\subsection{Baseline Methods}

We evaluated our approach against several baselines that are broadly categorized into 4 categories: 

\vspace{0.5em}
\noindent 1. End-to-End LLM QA: This approach leverages LLM for question answering without intermediate query structuring. The model receives a query and table as input, generating answers based on learned patterns and reasoning. We employ GPT-4o, GPT-4o-mini, Gemini-2.0-Flash, and DeepSeek-R1-Distill-Llama-70B for all tasks. 

\vspace{0.5em}
\noindent 2. Query Engines (Binder, BlendSQL): These methods generate hybrid SQL queries with LLM as \textit{User Defined Functions}. They leverage SQL to interpret tabular data and execute queries to retrieve relevant information. 

\vspace{0.5em}
\noindent 3. Pruning-Based Approach (ReacTable, H-STAR): These methods first apply SQL or Python-based pruning techniques, such as filtering columns or rows, before passing the refined table to an LLM for final answer extraction. 

\vspace{0.5em}
\noindent 4. Planning-Based Approach (ProTrix): ProTrix employs a two-step “Plan-then-Reason” framework. It first plans the reasoning, and assigns SQL to filter the table. Finally, it uses LLM to extract the final answer. By comparing our approach with these methods, we highlight the unique strengths of \methodName, which combines SQL-based filtering with LLM-driven reasoning for more effective query resolution.

\subsection{Results and Analysis}
\methodName performs well on three challenging benchmarks, we present key findings next.

\paragraph{First, are Hybrid Queries harder?} 

The results in Table \ref{tab:baseline_hybrid_compare} compare GPT-3.5-turbo on the original dataset with GPT-4o-mini on the hybrid set. Despite leveraging a more capable model (GPT-4o-mini) for the hybrid queries, we observe substantial performance drops, H-STAR and Binder see accuracy declines of 9.5\% and 32.7\%, respectively, on $\text{WikiTQ}_{\text{hybrid}}$. 

\begin{table}[h]
 \vspace{-0.75em}
\small
\setlength{\tabcolsep}{3.0pt}
\centering
\begin{tabular}{lcc}
\toprule
 & \shortstack{ Original (GPT-3.5)} & \shortstack{Hybrid (GPT-4o-mini)} \\
\midrule
\textit{Binder} & 56.7\% & 24\% \\
\textit{ReAcTable} & 52.4\% & 27\% \\
\textit{H-STAR} & 69.5\% & 59\% \\
\textit{ProTrix} & 65.2\% & 61.4\% \\
\bottomrule
\end{tabular}
\caption{\small Baselines result comparison on \textsc{WikiTQ} after filtering on hybrid part.}
\label{tab:baseline_hybrid_compare}
\vspace{-0.75em}
\end{table}

Notably, GPT-4o-mini outperforms GPT-3.5-turbo on benchmarks like MMLU and MATH (Source: OpenAI \footnote{\url{https://openai.com/index/gpt-4o-mini-advancing-cost-efficient-intelligence}}), yet still struggles on hybrid queries. This underscores their inherent difficulty and highlights the limitations of current methods in handling multi-step, semantically complex reasoning in Table QA.

\paragraph{Does \textit{\methodName} Help?}
Table \ref{tab:main_results} demonstrates that \methodName consistently outperforms state-of-the-art baselines across all datasets. On $\text{WikiTQ}_{\text{hybrid}}$, \methodName surpasses the best-performing baseline ProTrix by 5.5\% across all four models. On $\text{TabFact}_{\text{hybrid}}$, it achieves a breakthrough 91.2\% using DeepSeek model, surpassing the 90\% benchmark. On $\text{FinQA}_{\text{hybrid}}$, it achieves accuracy of 65\%, outperforming baselines by 4.6\% on DeepSeek-R1-Distill-Llama-70B.

\begin{table}[ht]
\small
\setlength{\tabcolsep}{3pt}
\centering
\begin{tabular}{lccc}
\toprule
 & \textbf{WikiTQ}$_{\text{hybrid}}$ & \textbf{TabFact}$_{\text{hybrid}}$ & \textbf{FinQA}$_{\text{hybrid}}$ \\
\midrule
& \multicolumn{3}{c}{GPT-4o-mini} \\
\midrule
\textit{End-to-End QA} & 60.4 & 84.4 & 44.7 \\
\textit{Binder$^{*}$} & 24.0 & 62.0 & 13.0 \\
\textit{BlendSQL} & 26.0 & 68.5 & 37.0 \\
\textit{ReAcTable$^{*}$} & 29.9 & 37.4 & - \\
\textit{H-STAR} & 59.0 & 83.0 & 40.1 \\
\textit{ProTrix} & 61.4 & 81.5 & 46.4 \\ 
\hline
\textbf{\textit{\methodName}} & \textbf{65.0} & \textbf{89.4} & \textbf{49.3} \\
\midrule
 & \multicolumn{3}{c}{GPT-4o} \\
\midrule
\textit{End-to-End QA} & 66.4 & 80.8 & 58.3 \\
\textit{Binder$^{*}$} & 27.3 & 60.3 & 17.0 \\
\textit{BlendSQL} & 42.0 & 68.3 & 34.3 \\
\textit{ReAcTable$^{*}$} & 45.4 & 45.4 & - \\
\textit{H-STAR} & 61.0 & \textbf{87.0} & 46.0 \\
\textit{ProTrix} & 61.7 & 80.5 & 54.3 \\ \hline
\textbf{\textit{\methodName }} & \textbf{70.7} & \underline{83.4} & \textbf{60.8} \\
\midrule
  & \multicolumn{3}{c}{Gemini-2.0-Flash} \\
\midrule
\textit{End-to-End QA} & 67.5 & 81.8 & 29.4 \\
\textit{Binder$^{*}$} & 12.9 & 60.4 & 21.3 \\
\textit{BlendSQL} & 31.1 & 60.1 & 19.7 \\
\textit{ReAcTable$^{*}$} & 20.4 & 37.6 & - \\
\textit{H-STAR} & 63.5 & \textbf{86.1} & 38.7 \\
\textit{ProTrix} & 62.2 & 80.8 & 42.9 \\ \hline
\textbf{\textit{\methodName}} & \textbf{69.6} & \underline{85.4} & \textbf{44.5} \\
\midrule
& \multicolumn{3}{c}{DeepSeek-R1-Distill-Llama-70B} \\
\midrule
\textit{End-to-End QA} & \textbf{76.4} & 82.5 & 52.4 \\
\textit{Binder$^{*}$} & 26.4 & 62.7 & 24.4 \\
\textit{BlendSQL} & 32.2 & 50.8 & 36.7 \\
\textit{ReAcTable$^{*}$} & 52.2 & 45.6 & - \\
\textit{H-STAR} & 68.7 & 55.6 & 50.3 \\
\textit{ProTrix} & 41.4 & 81.1 & 60.4 \\ \hline
\textbf{\textit{\methodName}} & \underline{73.0} & \textbf{91.2} & \textbf{65.0} \\
\bottomrule
\end{tabular}
\caption{\small Experimental results for various models on short-form QA, fact verification, and numerical reasoning tasks. $^{*}$: with self-consistency. Best result in \textbf{bold}, second-best in \underline{underlined}. A hyphen (-) indicates missing results due to incompatibility or untested scenarios.}
\vspace{-0.75em}
\label{tab:main_results}
\end{table}

\paragraph{{\methodName vs Query Engines:}} Binder and BlendSQL struggle with hybrid queries due to their rigid single-step execution framework. We observe that only 61\% and 66\% of the hybrid queries execute successfully in Binder and BlendSQL on $\text{WikiTQ}_{\text{hybrid}}$. The reported accuracies for these methods are calculated based on successfully executed queries. BlendSQL slightly outperforms Binder with a modest 2\%. \methodName outperforms BlendSQL by 39\%, 20.9\% and 12.3\% accuracy on $\text{WikiTQ}_{\text{hybrid}}$, $\text{TabFact}_{\text{hybrid}}$ and $\text{FinQA}_{\text{hybrid}}$ using GPT-4o-mini.

\paragraph{{\methodName vs Pruning Methods:}} H-STAR and ReAcTable while effective for structured queries, perform poorly on semantic tasks. H-STAR attains 59\% and 63.5\% accuracy on $\text{WikiTQ}_{\text{hybrid}}$ using GPT-4o-mini and Gemini-2.0-Flash, respectively, but struggles with row extraction. In some cases, its row-filtering heuristics discard critical contextual data essential for reasoning. H-STAR's higher accuracy on $\text{TabFact}_{\text{hybrid}}$ with GPT-4o stems from the dataset’s suitability for pruning techniques. \footnote{We did not test ReAcTable on $\text{FinQA}_{\text{hybrid}}$ since its prompts are tailored to other sets; modifying them changes baseline.} Furthermore, \textit{\methodName} with GPT-4o-mini and DeepSeek-R1-Distill-Llama-70B, despite their smaller size, performs competitively highlighting its effectiveness in resource-constrained environments where lightweight models are preferred. 

\paragraph{{\methodName vs Planning Method:}} ProTrix follows a two-step pipeline (planning and execution), achieves 61.4\% and 62.2\% on $\text{WikiTQ}_{\text{hybrid}}$ with GPT-4o-mini and Gemini-2.0-Flash. However, it fails in scenarios requiring intermediate semantic processing. For example, queries demanding dynamic column generation (e.g., inferring Country from driver name) reveal its inability to seamlessly integrate SQL and LLM reasoning, leading to a 9\% accuracy gap (GPT-4o) compared to \methodName.

\paragraph{{\methodName on Multimodal Data:}}  We evaluated \methodName on two multimodal datasets, FinQA$_{\text{MM}}$ and OTT-QA$_{\text{MM}}$, requiring multi-hop reasoning over both structured (tables) and unstructured (text) data. As shown in Table \ref{tab:text_table_results}, \methodName outperformed baselines, with notable improvements in FinQA$_{\text{MM}}$ and moderate gains in OTT-QA$_{\text{MM}}$.

\begin{table}[!htbp]
\vspace{-0.75em}
\small
\setlength{\tabcolsep}{0.5pt}
\centering
\begin{tabular}{lcccc}
 & FinQA$_{\text{MM}}$ & OTT-QA$_{\text{MM}}$  & FinQA$_{\text{MM}}$  & OTT-QA$_{\text{MM}}$  \\
\toprule
 & \multicolumn{2}{c}{GPT-4o-mini} & \multicolumn{2}{c}{GPT-4o} \\
\textit{End2End QA} & 45.9 & 61.2 & 57.6 & \textbf{68.7}  \\
\textbf{\textit{\methodName}} & \textbf{63.2} & \textbf{63.7} & \textbf{68.0} &  65.2 \\
\toprule
 & \multicolumn{2}{c}{Gemini-2.0-Flash} & \multicolumn{2}{c}{DeepSeek-R1$^{*}$} \\
\midrule
\textit{End2End QA} & 37.9 & 64.1 & 54.8 &  59.9 \\
\textbf{\textit{\methodName}} & \textbf{60.8} & \textbf{66.7}  & \textbf{66.2} & \textbf{62.8} \\
\bottomrule
\end{tabular}
\caption{\small Experimental results on multimodal QA (tables + paragraphs). $^{*}$: DeepSeek-R1 refers to DeepSeek-R1-Distill-Llama-70B.}
\label{tab:text_table_results}
\vspace{-0.75em}
\end{table}

In FinQA$_{\text{MM}}$, \methodName excelled in numerical and multi-hop reasoning, where end-to-end models struggled with irrelevant information and sequential logic, such as calculating net values. For OTT-QA$_{\text{MM}}$, which involved less structured computation and more real-world knowledge, \methodName still showed consistent gains. Unlike baselines, \methodName effectively retrieved and integrated key table and paragraph segments, ensuring relevant information was used. These results highlight the strength of our modular, reasoning-focused approach, which integrates information step-by-step instead of relying on holistic attention.

On MMTabQA$_{\text{MM}}$ dataset, \methodName achieved an accuracy of 53.02\% using gpt-4o-mini model, significantly outperforming the end-to-end QA baseline, which scored 46.33\%. These results highlight the strength of our modular, reasoning-driven approach, combining structured data (tables), unstructured data (text), and visual inputs (images) for superior performance. This underscores \methodName's versatility and scalability in addressing complex multimodal (table, text, images) QA tasks.

\paragraph{Efficacy Analysis:}

We conducted an analysis to assess the effectiveness and scalability of \methodName in $\text{WikiTQ}_{\text{hybrid}}$, focusing on 98 large tables with over 30 rows and average token length of  17,731 per table. Our results show that \methodName achieved an accuracy of 65.6\%, outperforming ProTrix (37.5\%) and H-STAR (35.9\%) by 28.1\% and 29.7\%, respectively. On these large tables, \methodName achieved 65.6\%, the same accuracy as on the entire dataset, demonstrating its ability to handle complex queries while remaining both scalable and robust. These results highlight \methodName’s ability to tackle a wide range of table-based tasks with consistent performance.

In addition to delivering reliable performance, \textit{\methodName} offers the critical advantage of transparent, interpretable reasoning. By following a structured execution plan, it ensures that final answers are tightly aligned with preceding reasoning steps, enhancing traceability and reducing spurious outputs. This directly addresses a core limitation of large language models, hallucination and memorization. Unlike end-to-end LLM, which may produce correct answers without valid reasoning, \textit{\methodName} only yields correct outputs when the underlying plan is sound, ensuring both reliability and interpretability.

\paragraph{Efficiency Analysis:}
Table \ref{tab:api_calls} demonstrate the efficiency of our proposed \methodName framework using number of API calls. We make about six API calls which are much lower compared to approaches that use self-consistency (Binder) with 50 calls and H-STAR which uses $\sim$ 8 calls to reach the answer. 

\begin{table}[!htbp]
\small
\setlength{\tabcolsep}{3pt}
	\centering
	\begin{tabular}{lccc}
		\toprule
API calls/ Query & GPT-4o & GPT-4o-mini & Gemini-2.0$^{*}$\\
 \midrule

  \textit{ProTrix} & 2&2 & 2 \\
  \textit{Binder} &50& 60 & 53  \\
  \textit{H-STAR} & 8&8& 8\\
  \textbf{\textit{\methodName}} &5.31 & 5.87 & 5.85  \\
    \bottomrule
	\end{tabular}
	\caption{\small Number of API calls comparison per Table QA. $^{*}$: Gemini-2.0 refers to Gemini-2.0-Flash.}
 \vspace{-1.0em}
	\label{tab:api_calls}
\end{table}

ProTrix uses only two fixed API calls and relies on the LLM solely for generating the plan. However, it does not involve the LLM during execution. This limits its ability to handle multi-step queries requiring reasoning at each step. For instance, it may fail to infer information from individual rows or perform operations such as applying a SQL GROUP BY on an LLM-inferred \texttt{country} column. Such steps are often essential to arrive at the correct final answer. These metrics demonstrate how our approach minimizes computational overhead while maintaining accuracy.

\paragraph{Optimization:} We experimented with optimizing our planning and execution strategy on 200 Table QA queries, which were randomly sampled from the evaluation benchmarks. Table \ref{tab:optimization_results} demonstrates how our optimization strategy focuses on reducing unnecessary computational steps without compromising accuracy. 
\begin{table}[H]
\vspace{-0.75em}
\small
\setlength{\tabcolsep}{3pt}
\centering
\begin{tabular}{l c c}
\toprule
\multicolumn{3}{c}{\#LLM Optimization Effect} \\
\midrule
\text{\#LLM Drops} & \multicolumn{2}{c}{15} \\
\text{\#SQL Drops} & \multicolumn{2}{c}{19} \\
\text{\#SQL Merge} & \multicolumn{2}{c}{113} \\
\text{\#SQL Reorder} & \multicolumn{2}{c}{4} \\
\midrule
 & \textbf{Before Opt.} & \textbf{After Opt.} \\
\midrule
\text{\#LLM} & 74 & 59 \\
\text{\#SQL} & 532 & 513 \\ 
\text{\#Total Steps} & 616 & 469 \\
\text{Accuracy (\%)} & 65 & 64 \\
\bottomrule
\end{tabular}
\caption{\small Effect of optimization on GPT-4o-mini plans on \textsc{WikiTQ}. Opt. stands for plan optimization.}
\label{tab:optimization_results}
\vspace{-1.0em}
\end{table}    
This optimization was achieved by targeting redundancies in both the LLM and SQL steps.

\paragraph{1. {LLM Step Reduction}:} By identifying and eliminating redundant LLM steps, we reduced the 15 LLM calls. This optimization ensures that LLMs are only used when necessary, lowering computational costs.

\paragraph{2. {SQL Step Optimization}:} We achieved a reduction of 19 SQL steps by eliminating unused operations. Additionally, we merged 113 sequential SQL steps into fewer, more efficient queries and reordered 4 steps to optimize execution flow, making it more efficient.

Optimizing GPT-4o-mini reduces LLM steps by 20\% and SQL steps by 24.8\%, significantly improving efficiency. The accuracy slightly drops from 65\% to 64\%, but this trade-off is minimal, especially under practical constraints. Our goal is to create a scalable Table QA pipeline that balances accuracy with computational cost, particularly for large tables. The optimization achieves this by maintaining modular reasoning while reducing latency, API calls, and input tokens, demonstrating that efficiency gains don’t sacrifice performance.

\begin{table}[H]
\vspace{-0.5em}
\small
\setlength{\tabcolsep}{3pt}
\centering
\begin{tabular}{lccc}
\toprule
 & GPT-4o & GPT-4o-mini & Gemini-2.0$^{*}$ \\
 \midrule
 & \multicolumn{3}{c}{\textit{ProTrix}} \\ \midrule
\textit{SQL Error} & 51.2\% & 25.9\% & 27.0\%\\
\textit{Plan Generation} & 11.0\% & 17.0\% & 9.0\% \\
\toprule
 & \multicolumn{3}{c}{\textbf{\textit{\methodName}}} \\
\midrule
\textit{SQL Error} & 15.0\% & 42.5\% & 16.0\%  \\
\textit{Plan Generation} & 1.0\% & 3.0\% & 1.0\% \\
\bottomrule
\end{tabular}
\caption{\small Error in SQL and Plan Generation on \textsc{WikiTQ}. $^{*}$: Gemini-2.0 refers to Gemini-2.0-Flash.}
\label{tab:error_analysis}
\vspace{-0.75em}
\end{table}
\paragraph{Error Analysis:}
Table~\ref{tab:error_analysis} illustrates the effectiveness of our approach in minimizing errors in SQL execution and plan generation. Our approach reduces SQL errors by 30\% and plan generation by 86\% compared to ProTrix in both GPT-4o, GPT-4o-mini and Gemini-2.0-Flash. However, GPT-4o-mini exhibits a higher SQL error rate due to its smaller model size, which limits its ability to generate accurate SQL queries.

In \methodName, SQL errors arise due to incorrect formatting, unsupported MySQL functions, or hallucinated columns and tables. Planning errors arise when SQL steps replace LLM reasoning or generate unused tables. The Plan Verification Step, Figure \ref{fig:main_figure}, mitigates these issues by refining planning for improved reliability in complex table-based reasoning.

\paragraph{Analysis Across Pipeline Stages:}
We perform a stage-wise analysis to assess the contribution of each component in our pipeline: \textit{filtering}, \textit{planning}, and \textit{execution}. Compared to SQL-only generation, which struggles with multi-step reasoning, our pipeline yields consistent accuracy gains by structuring the task into sub-components. The \textit{filtering} stage removes irrelevant columns, reducing noise and guiding the model’s attention. The \textit{planning} stage, central to our method, decomposes complex queries into symbolic and semantic steps. This step is essential and cannot be ablated. However, skipping plan verification (i.e., executing without validating) leads to a 1\% drop in accuracy. Finally, \textit{execution} stage translates structured plans into SQL query and LLM prompts with column details.

\section{Comparison with Related Work}
Table-based Question Answering (Table QA) combines table understanding, question interpretation, and NLP. Foundational work such as Text-to-SQL \cite{text2sql}, Program-of-Thought \cite{pot}, and TabSQLify \cite{tabsqlify} laid the groundwork. Binder and TAG \cite{TAG} expose the limitations of traditional Text-to-SQL methods in handling complex analytical tasks involving both structured and unstructured data. To address these challenges, several alternative approaches have been explored:

\paragraph{Fine-Tuning Methods:}
These methods fine-tune LLMs to specialize in reasoning over hybrid tabular and textual data. Models such as \cite{TAT}, \cite{ssql}, and \cite{LOTUS} are trained to extract, reason, and execute over such inputs. However, fine-tuning requires large task-specific datasets and tends to lack generalization across domains.

\paragraph{Query Engines:}
This direction integrates LLM with SQL engines via user-defined functions (UDFs), allowing LLM calls within queries. UQE \cite{UQE}, BlendSQL, SUQL \cite{SUQL}, and Binder follow this paradigm. While flexible, LLM-generated queries can be error-prone, and these systems often support only limited query structures, reducing adaptability.

\paragraph{Table Pruning and Planning:}
Approaches like H-STAR, ReAcTable, ProTrix, and others \cite{relational_workload, reasoning} enhance efficiency by programmatically pruning rows or columns using SQL or Python. While this reduces processing overhead, these methods often function as black boxes, lacking transparency and vulnerable to cascading errors if early pruning steps are incorrect. More information in Appendix \ref{appeddix:comparison_with_method}.

\section{Conclusion}
We introduce \methodName, a novel approach for table-based question answering on tables with embedded unstructured text. \methodName outperforms all baselines by strategically decomposing complex queries into a sequence of LLM- and SQL-based planning steps. By alternating between these modalities, it enables precise, interpretable, and adaptive query resolution. \methodName overcomes prior limitations by effectively handling both complex queries and large tables. Its modular design also supports future extensions, including image-based tables, multi-table reasoning, and integration with free-form text. As future work, we plan to explore fine-tuning and supervision strategies to further improve execution accuracy and plan reliability.

\section*{Limitations}
While our approach demonstrates strong performance across multiple datasets, it is currently limited to English-language tables, restricting its applicability to multilingual settings. Additionally, our method does not explicitly handle hierarchical tables, where multi-level dependencies introduce additional complexity in reasoning. Another limitation is the inability to process multi-table queries, which require reasoning across multiple relational structures. Furthermore, the lack of well-established benchmarks for hybrid datasets poses a challenge in evaluating and further improving performance in more complex, real-world scenarios. 

\section*{Ethics Statement}
We, the authors, confirm that our research adheres to the highest ethical standards in both research and publication. We have thoughtfully addressed various ethical considerations to ensure the responsible and equitable use of computational linguistics methodologies. In the interest of reproducibility, we provide detailed resources, including publicly available code, datasets (compliant with their respective ethical standards), and other relevant materials. Our claims are supported by the experimental results, although some degree of stochasticity is inherent in black-box large language models, which we mitigate by using a fixed temperature. We also offer thorough details on annotations, dataset splits, models used, and prompting techniques to ensure that our work can be reliably reproduced. We used AI assistants to help refine the writing and improve clarity during the drafting and revision process. No content was generated without human oversight or verification.

\section*{Acknowledgments}
We gratefully acknowledge the Cognitive Computation Group at the University of Pennsylvania and the Complex Data Analysis and Reasoning Lab at Arizona State University for their resources and computational support. A part of this work was funded by ONR Contract N00014-23-1-2364 and N00014-23-1-2417. Yanjie is supported by the National Science Foundation (NSF) via the grant numbers: 2426340, 2416727, 2421864, 2421865, 2421803, and National academy of engineering Grainger Foundation Frontiers of Engineering Grants. We also thanks the reviewers for there thoughtful feedback and comment.

\bibliography{anthology,custom}
\bibliographystyle{acl_natbib}

\appendix

\newpage
\appendix
\onecolumn
\section{Appendix: LLM Prompts and Examples}
\label{appendix:prompt_and_examples}
\subsection{Prompt Examples}

\begin{lstlisting}[language=Prompt,caption={Extract Relevant Column}]
Given column descriptions, Table and Question return a list of columns that can be
relevant to the solving the question (even if slightly relevant) given the table
name and table:
table name: {self.name}

table: {self.table}
Question: {self.question}

Example output: [ 'Score', 'Driver']
Instructions:
1. Do not provide any explanations, just give the cols as a list
2. The list will be used to filter the table dataframe directly so take care of that

Output:
\end{lstlisting}

\begin{lstlisting}[language=Prompt,caption={Column Description Prompt}]
Give me the column name, data type, formatting that needs to be done, column 
descriptions in short for the given table and question. The descriptions should be 
useful in planning steps that solve the question asked on that table. Also, give a 
small description of the table using table name and table data given.
Table:
table name: {self.name}
{self.table}
Question: {self.question}
\end{lstlisting}

\begin{lstlisting}[language=Prompt,caption={Planning Prompt}]
I need a step-by-step plan in plain text for solving a question, given column
descriptions and table rows. Follow these guidelines:
Begin analyzing the question to categorize tasks that require only SQL capabilities 
(like straightforward data formatting, mathematical operations, basic aggregations) 
and those that need LLM assistance (like summarization, text interpretation, or 
answering open-ended queries).
MySQL Query Generation: For parts of the question that involve formatting of column
data type, filtering and mathematical or analytical tasks, generate SQL query code 
to answer them directly, without using an LLM call.
LLM-Dependent Task Identification: For tasks that SQL cannot inherently 
perform, specify the columns or portions of rows where LLM calls are needed. Add an 
extra column in the result set to store the LLM output for each row in the filtered 
data subset.
Example -
<Table Name>
<Table>
Question: <Question>
<Column Descriptions>
<Plan>

Solve for this:
Table:
table name: {self.name}
{self.table}
Question: {self.question}
{self.description}
Only give the step-by-step plan and remove any other explanations or code.
Output format:
Step 1: SQL
Step 2: Either SQL or LLM
Step 3: ...
Step 4: ...
\end{lstlisting}

\begin{lstlisting}[language=Prompt,caption={Verify Plan Prompt}]
Suppose you are an expert planner verification agent. 
Verify if the given plan will be able to answer the Question asked on this table.
Table name: {self.name}
Table: {self.table}
Column descriptions: {self.description}
Question to Answer: {self.question}
Old Plan: {self.plan}
Is the given plan correct to answer the Question asked on this table (check format
issues and reasoning steps) should be able to guide the LLM to write correct code
and get correct result.
If the plan is not correct, provide better plan detailed on what needs to be done
handling all kinds of values in the column.
- Check if the MySQL step logic adheres to the column format. (Performs calculations 
and formatting and filtering in the table)
- The LLM step's logic will help in getting the correct answer.
If the original plan is correct then return that plan.

Do not provide code or other explanations, only the new plan.
Output format:
Step 1: Either SQL or LLM - ...
Step 2: SQL or LLM - ...
Step 3: SQL ...

As given in original plan.
\end{lstlisting}

\begin{lstlisting}[language=Prompt,caption={Code Execution Prompt}]
MySQL_Code_Generation: For parts of the question that involve data
formatting, data manipulations such as filtering, grouping, aggregations, and
creating new tables. Generate optimized MySQL code to 
answer those parts directly without using an LLM.

LLM-Dependent Tasks Identification: For tasks that SQL cannot inherently perform
that require sentiment analysis, logical inferences, or questions that involve 
interpreting text data, specify only that 1 column where LLM calls are needed. Add 
an extra column in the table that stores the LLM output for each row in the filtered 
data subset.

Instructions:
1. Store the output at each step by creating a new table. Use this new table for the 
next steps.
2. The code for MySQL should handle all values in the column (formatting and
filtering). New columns 
from previous LLM steps can be assumed present in table.
3. Don't give any other explanations, only MySQL and LLM steps as the given plan.

Then, Only give step (SQL or LLM) that is needed -
The Output format example -
Step_1 - SQL: MySQL code, table name to be used in the next query
Step_2 - LLM:
- Reason: Why we need to use LLM
- Table name:
- original column to be used:
- LLM prompt: The prompt that user can use to solve the problem
- New column name:
Step_3 - SQL: MySQL code, table name to be used in the next query
Step_4 - ...
Step_5 - ...

LLM step format should be the same.
Solve for this question, given table and step by step plan as a reference:
Table name: {self.name}
Schema: {self.table.columns}
Column Descriptions: {self.description}
Table: {self.table}
Question: {self.question}
Plan: {self.plan}

First check if taking above Plan will give the desired output.
Give me code for solving the question, and no other explanations. Keep in mind the
column data formats (string to appropriate data type, removing extra character, Null
values) while writing Mysql code.
\end{lstlisting}

\begin{lstlisting}[language=Prompt,caption={LLM Step Prompt}]
Given a column and step you need to perform on it -
Column: {df.column}
Step to solve the question: {step.prompt}
Question: {self.question}

Instructions:
- Do not provide any explanation and Return only a list (separate values by '#')
that can be added 
to a dataframe as a new column in a dataframe.
- Do not create a column name already present in the table. (duplicate column)
- Any value should not be more than 3 words (or each value should be as short as
possible).
- Size of output list Should be same as input list.
\end{lstlisting}

\begin{lstlisting}[language=Prompt,caption={Answer Extraction Prompt}]
Table: {self.name}
{self.table}
Question: {self.question}

Answer the question given the table in as short as possible.
If the table has just one column or value consider that as the answer given the 
column name.
Just provide the answer, do not provide any other information.
\end{lstlisting}

\begin{lstlisting}[language=Prompt,caption={Answer Format Prompt}]
Table: {self.name}
{self.table}
Question: {self.question}

You will be given Answer and Gold Answer, you have to Convert the answer into a 
format of gold answer given above, if the content or meaning is same (semantically 
same) they should be same.

Few examples of conversion for your understanding:
1. answer: ITA, gold answer: Italy. Reasoning- ITA is country code of Italy hence ITA and Italy are same and you can convert ITA to Italy.
    Your Output: Italy
2. answer: 17, gold answer: 17 years. Reasoning- 17 of answer is same as 17 years of the gold answer in the given context of question.
    Your Output: 17 years
3. answer : 10, gold answer: 10. Reasoning- Since, both values are already same no conversion is needed.
    Your Output: 10
4. answer : 0, gold answer: 5. Reasoning- Since, both values are semantically not same no conversion is needed for the answer.
    Your Output: 0
5. answer : The answer is not present in the table. , gold answer: 5. Reasoning- Since, both values are semantically not same no convertion is needed for the answer.
    Your Output: The answer is not present in the table.
\end{lstlisting}

\begin{lstlisting}[language=Prompt,caption={Plan Optimization Prompt}]
You are an expert in SQL and plan optimization. Your task is to optimize the given 
SQL plan while ensuring it correctly answers the given question. Use the following 
optimization strategies, but only if they maintain correctness:

SQL Merge: Merge sequential SQL steps where possible (e.g., combining filtering,
aggregation, and sorting in one query).
SQL Reordering: Reorder SQL steps to filter early before applying computationally
expensive operations like LLM processing.
LLM Merge: Merge sequential LLM steps where the operation is on the same column.
Given Information: ...
Plan: {self.plan}
\end{lstlisting}

\subsection{Table and Question Example}
\label{appendix:exampleTqbleQA}
Below is an example of a table and question used for LLM planning.

\begin{lstlisting}[language=Prompt,caption={Table and Question Example for LLM Planning}]
Table: New_York_Americans_soccer
           Year  Division League        Reg_Season               Playoffs National_Cup
0          1931       1.0    ASL        6th (Fall)             No playoff         None
1   Spring 1932       1.0    ASL              5th?             No playoff    1st Round
2     Fall 1932       1.0    ASL               3rd             No playoff         None
3   Spring 1933       1.0    ASL                 ?                      ?        Final
4       1933/34       NaN    ASL               2nd             No playoff            ?
5       1934/35       NaN    ASL               2nd             No playoff            ?
6       1935/36       NaN    ASL               1st  Champion (no playoff)            ?
7       1936/37       NaN    ASL     5th, National        Did not qualify     Champion
8       1937/38       NaN    ASL  3rd(t), National              1st Round            ?
9       1938/39       NaN    ASL     4th, National        Did not qualify            ?
10      1939/40       NaN    ASL               4th             No playoff            ?
11      1940/41       NaN    ASL               6th             No playoff            ?
12      1941/42       NaN    ASL               3rd             No playoff            ?
13      1942/43       NaN    ASL               6th             No playoff            ?
14      1943/44       NaN    ASL               9th             No playoff            ?
15      1944/45       NaN    ASL               9th             No playoff            ?
16      1945/46       NaN    ASL               5th             No playoff            ?
17      1946/47       NaN    ASL               6th             No playoff            ?
18      1947/48       NaN    ASL               6th             No playoff            ?
19      1948/49       NaN    ASL            1st(t)                 Finals            ?
20      1949/50       NaN    ASL               3rd             No playoff            ?
21      1950/51       NaN    ASL               5th             No playoff            ?
22      1951/52       NaN    ASL               6th             No playoff            ?
23      1952/53       NaN    ASL               6th             No playoff   Semifinals
24      1953/54       NaN    ASL               1st  Champion (no playoff)     Champion
25      1954/55       NaN    ASL               8th             No playoff            ?
26      1955/56       NaN    ASL               6th             No playoff            ?

Question_: How long did it take for the New York Americans to win the National Cup
after 1936?
\end{lstlisting}

\subsection{Model Responses}
\begin{lstlisting}[language=Prompt,caption={Response to Sample Prompt 1 (Column Description)}]
### Table Description
The table `New_York_Americans_soccer` contains historical performance data for the
New York Americans soccer team, detailing their standings in various leagues, 
playoff outcomes, and national cup results from 1931 to 1956. The data includes
information on the year, division, league, regular season standings, playoff 
participation, and national cup results.

### Column Details

| Column Name        | Data Type   | Formatting Needed                     | Column Description                                                                 |
|--------------------|-------------|------------------------|--------------|
| Year               | String      | Standardize to a consistent format    
| Represents the year or season of the soccer performance. Some entries may need parsing to extract the year. |
| Division           | Float       | Convert to Integer (if applicable)    
| Indicates the division in which the team played. Some entries are NaN and need handling. |
| League             | String      | Standardize to a consistent format    
| Represents the league in which the team participated, primarily ASL (American Soccer League). |
| Reg_Season         | String      | Clean and standardize standings 
| Indicates the team's regular season performance, which may include qualifiers like "t" or "?" that need clarification. |
| Playoffs           | String      | Standardize and clean           
| Indicates playoff participation and outcomes, with entries like "No playoff" and "Finals" that need to be standardized. |
| National_Cup       | String      | Standardize and clean
| Indicates the outcome of the national cup, with entries like "Champion" or "1st Round" that need standardization. |

\end{lstlisting}

\begin{lstlisting}[language=Prompt,caption={Response to Sample Prompt 2 (Planning Prompt)}]
Plan: Step 1: SQL - Standardize the Year column to a consistent format and extract
the year from entries like "Spring 1932" and "Fall 1932".
Step 2: SQL - Clean and standardize the National_Cup column to identify the years
when the team won the national cup.
Step 3: SQL - Filter the data to find the first year after 1936 when the 
National_Cup column indicates a win.
\end{lstlisting}

\begin{lstlisting}[language=Prompt,caption={Response to Sample Prompt 3 (Verify Plan Prompt)}]
New Plan: ### Revised Plan:
Step 1: LLM - Standardize the Year column to a consistent format by extracting the
year from entries like "Spring 1932" and "Fall 1932". Convert all entries to a four-
digit year format (e.g., "1932" instead of "Spring 1932").
Step 2: SQL - Clean and standardize the National_Cup column to identify winning 
entries. Define a clear criterion for a "win," such as entries that contain 
"Champion" or "1st Round" (if applicable).
Step 3: SQL - Filter the data to find the first year after 1936 where the 
National_Cup column indicates a win. Ensure to handle any NaN or ambiguous entries 
appropriately.
\end{lstlisting}

\subsection{Code Execution Results}
\begin{lstlisting}[language=Prompt,caption={Response to Sample Prompt 4 (Code Execution Prompt) Step 1}]
LLM_Step - 
- Reason: Standardize the Year column to correct format.
- Table name: New_York_Americans_soccer
- original column to be used: Year
- LLM prompt: Extract the year from phrases like "Spring 1932" or "Fall 1932" and
standardize all entries to a YYYY format. Ensure the output is consistent across all 
entries.
- New column name: Year_Formatted.

Dataframe create after LLM Step

Year_Formatted Division League    Reg_Season              Playoffs           National_Cup
0   1931       1.0      ASL        6th (Fall)             No playoff           None
1   1932       1.0      ASL              5th?             No playoff      1st Round
2   1932       1.0      ASL               3rd             No playoff           None
3   1933       1.0      ASL                 ?                     ?          Final
4   1933       NaN      ASL               2nd             No playoff              ?
5   1934       NaN      ASL               2nd             No playoff              ?
6   1935       NaN      ASL               1st  Champion (no playoff)              ?
7   1936       NaN      ASL     5th, National        Did not qualify       Champion
8   1937       NaN      ASL  3rd(t), National              1st Round              ?
9   1938       NaN      ASL     4th, National        Did not qualify              ?
10  1939       NaN      ASL               4th             No playoff              ?
11  1940       NaN      ASL               6th             No playoff              ?
12  1941       NaN      ASL               3rd             No playoff              ?
13  1942       NaN      ASL               6th             No playoff              ?
14  1943       NaN      ASL               9th             No playoff              ?
15  1944       NaN      ASL               9th             No playoff              ?
16  1945       NaN      ASL               5th             No playoff              ?
17  1946       NaN      ASL               6th             No playoff              ?
18  1947       NaN      ASL               6th             No playoff              ?
19  1948       NaN      ASL            1st(t)                 Finals              ?
20  1949       NaN      ASL               3rd             No playoff              ?
21  1950       NaN      ASL               5th             No playoff              ?
22  1951       NaN      ASL               6th             No playoff              ?
23  1952       NaN      ASL               6th             No playoff     Semifinals
24  1953       NaN      ASL               1st  Champion (no playoff)       Champion
25  1954       NaN      ASL               8th             No playoff              ?
26  1955       NaN      ASL               6th             No playoff              ?

\end{lstlisting}

\begin{lstlisting}[language=Prompt,caption={Response to Sample Prompt 4 (Code Execution Prompt) Step 2}]
SQL_Step - 
CREATE TABLE standardized_national_cup AS
SELECT 
    Year_Formatted,
    Division,
    League,
    Reg_Season,
    Playoffs,
    CASE 
        WHEN National_Cup LIKE Champion_ THEN Win__
        WHEN National_Cup LIKE Round__ THEN Win__
        ELSE No_Win
    END AS National_Cup
FROM New_York_Americans_soccer;
Table created: standardized_national_cup

Dataframe created after loading from standardized_national_cup ...
    Year_Formatted  Division League        Reg_Season               Playoffs National_Cup
0   1931       1.0    ASL        6th (Fall)             No playoff       No Win
1   1932       1.0    ASL              5th?             No playoff          Win
2   1932       1.0    ASL               3rd             No playoff       No Win
3   1933       1.0    ASL                 ?                      ?       No Win
4   1933       NaN    ASL               2nd             No playoff       No Win
5   1934       NaN    ASL               2nd             No playoff       No Win
6   1935       NaN    ASL               1st  Champion (no playoff)       No Win
7   1936       NaN    ASL     5th, National        Did not qualify          Win
8   1937       NaN    ASL  3rd(t), National              1st Round       No Win
9   1938       NaN    ASL     4th, National        Did not qualify       No Win
10  1939       NaN    ASL               4th             No playoff       No Win
11  1940       NaN    ASL               6th             No playoff       No Win
12  1941       NaN    ASL               3rd             No playoff       No Win
13  1942       NaN    ASL               6th             No playoff       No Win
14  1943       NaN    ASL               9th             No playoff       No Win
15  1944       NaN    ASL               9th             No playoff       No Win
16  1945       NaN    ASL               5th             No playoff       No Win
17  1946       NaN    ASL               6th             No playoff       No Win
18  1947       NaN    ASL               6th             No playoff       No Win
19  1948       NaN    ASL            1st(t)                 Finals       No Win
20  1949       NaN    ASL               3rd             No playoff       No Win
21  1950       NaN    ASL               5th             No playoff       No Win
22  1951       NaN    ASL               6th             No playoff       No Win
23  1952       NaN    ASL               6th             No playoff       No Win
24  1953       NaN    ASL               1st  Champion (no playoff)          Win
25  1954       NaN    ASL               8th             No playoff       No Win
26  1955       NaN    ASL               6th             No playoff       No Win

\end{lstlisting}

\begin{lstlisting}[language=Prompt,caption={Response to Sample Prompt 4 (Code Execution Prompt) Step 3}]
SQL_Step -
CREATE TABLE first_win_after_1936 AS
SELECT 
    Year_Formatted,
    Division,
    League,
    Reg_Season,
    Playoffs,
    National_Cup
FROM standardized_national_cup
WHERE Year_Formatted > 1936_ AND National_Cup = Win__
ORDER BY Year
LIMIT 1;
Table created: first_win_after_1936
Dataframe created after loading from first_win_after_1936 ...
   Year_Formatted Division League Reg_Season               Playoffs National_Cup
0  1953     None    ASL        1st  Champion (no playoff)          Win

\end{lstlisting}

\begin{lstlisting}[language=Prompt,caption={Response to Sample Prompt 5 (Answer Extraction Prompt)}]
Generated Answer: 17 years
Comparison Result: Yes
Actual answer: 17 years, model answer: 17 years
Answer matched: True
\end{lstlisting}

\subsection{More Detail on Optimization}
\label{appendix:optmization}
\begin{figure}[h!]
    \centering
  \includegraphics[width=0.48\textwidth]{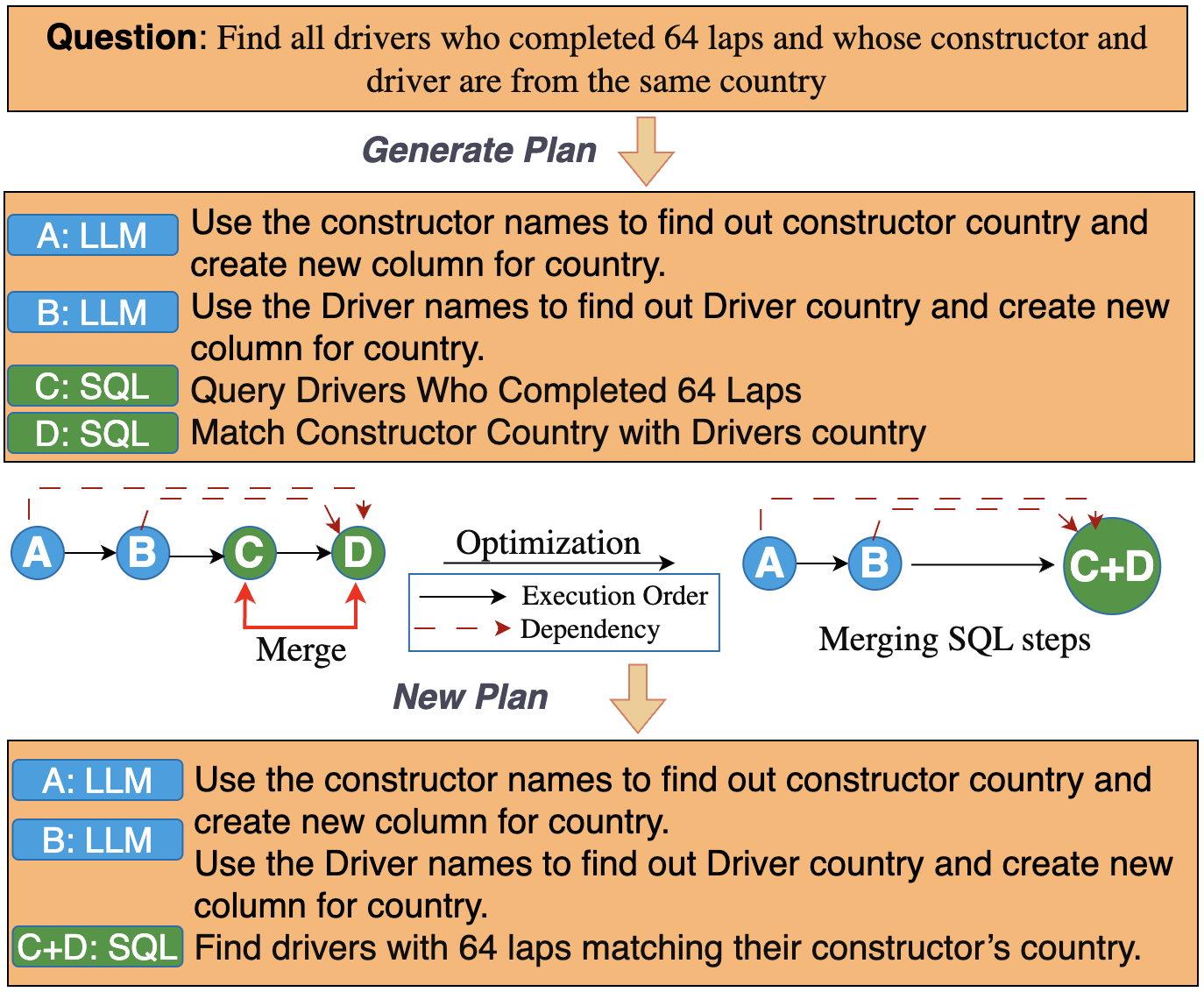}
    \caption{Optimization using SQL step merging}
    \label{fig:3}
\end{figure}

We have explored several optimization techniques to enhance the efficiency of our pipeline by reducing the number of steps generated during query execution. While some of these strategies are detailed in the section \ref{section:our-approach}, we outline additional key techniques below:

\paragraph{SQL Merging:} Figure \ref{fig:3} explains merging sequential SQL steps to optimize the pipeline’s performance. Since SQL operations follow a logical structure, combining multiple steps into a single query does not compromise the correctness or integrity of the process. This consolidation reduces the overhead of executing individual queries and improves the overall efficiency of the pipeline by minimizing redundant operations and streamlining execution.

\begin{lstlisting}[language=Prompt,caption={Example of Step Merging}]
Original Plan (Multiple SQL Steps):
SELECT * FROM table WHERE column = 'X';  
SELECT * FROM table ORDER BY date DESC;  

Optimized (Merged into a Single Step):
SELECT * FROM table WHERE column = 'X' ORDER BY date DESC; 

\end{lstlisting}

\paragraph{Parallel LLM Execution:}
Initially, we prompted the LLM to generate a new column by supplying the entire existing column and asking it to return a list of the same length. However, this approach often led to inconsistent results, such as incorrect list lengths, duplicated values, or hallucinated entries, due to the model’s sensitivity to long input sequences. Errors were especially prevalent in the middle of the list, consistent with the “Lost in the Middle” effect  \cite{lostinthemiddle}.

\begin{figure}[h!]
    \centering
\includegraphics[width=0.8\textwidth]{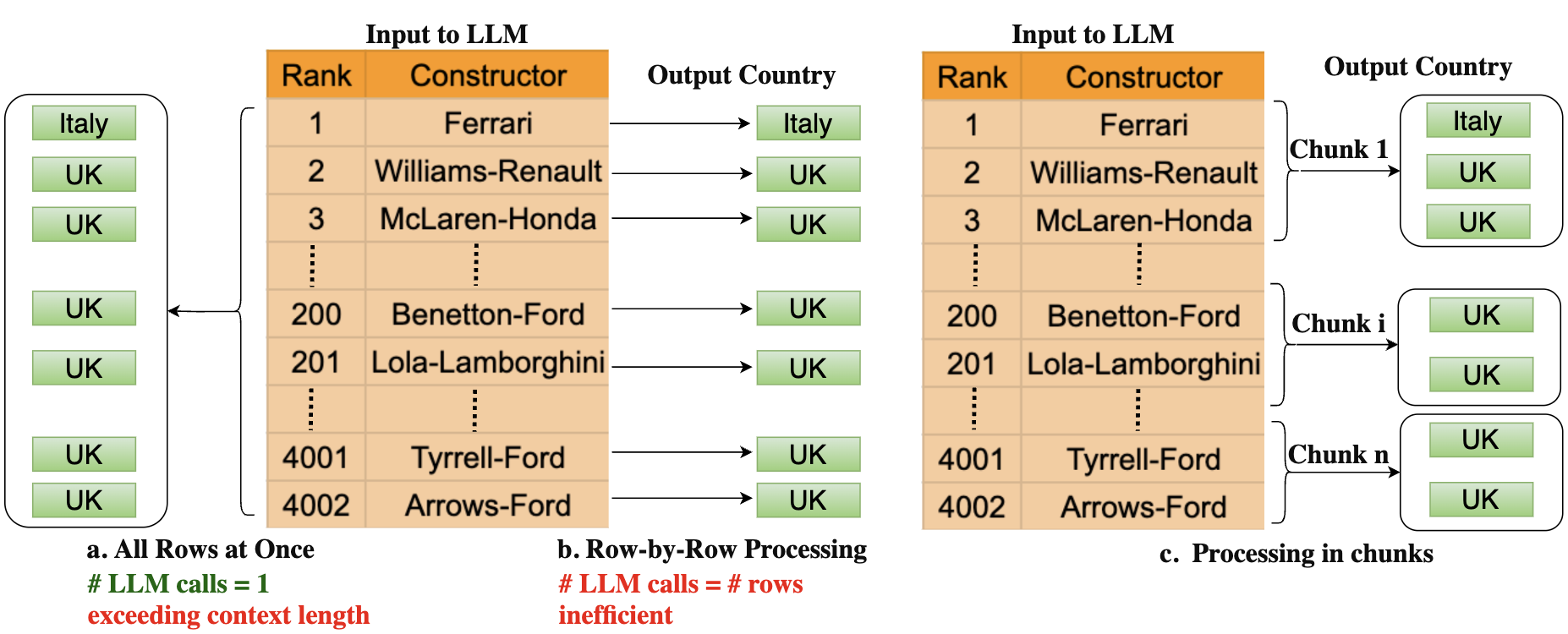}
    \caption{\small Optimizing LLM Calls: Chunk-based processing on Rows}
    \label{fig:llm-optimizaton-rows-batch}
\end{figure}

To improve both reliability and efficiency, we adopted a chunk-wise parallel execution strategy that avoids the overhead of row-by-row processing while enhancing consistency. As illustrated in Figure \ref{fig:llm-optimizaton-rows-batch}, we segment the input into appropriately sized batches and execute multiple LLM calls in parallel. This design enables simultaneous inference over different parts of the data, substantially reducing latency by eliminating sequential processing bottlenecks. The result is faster response times and improved scalability, making this approach well-suited for large-scale reasoning tasks over semi-structured data.

\begin{figure}[H]
    \centering
  \includegraphics[width=0.6\textwidth]{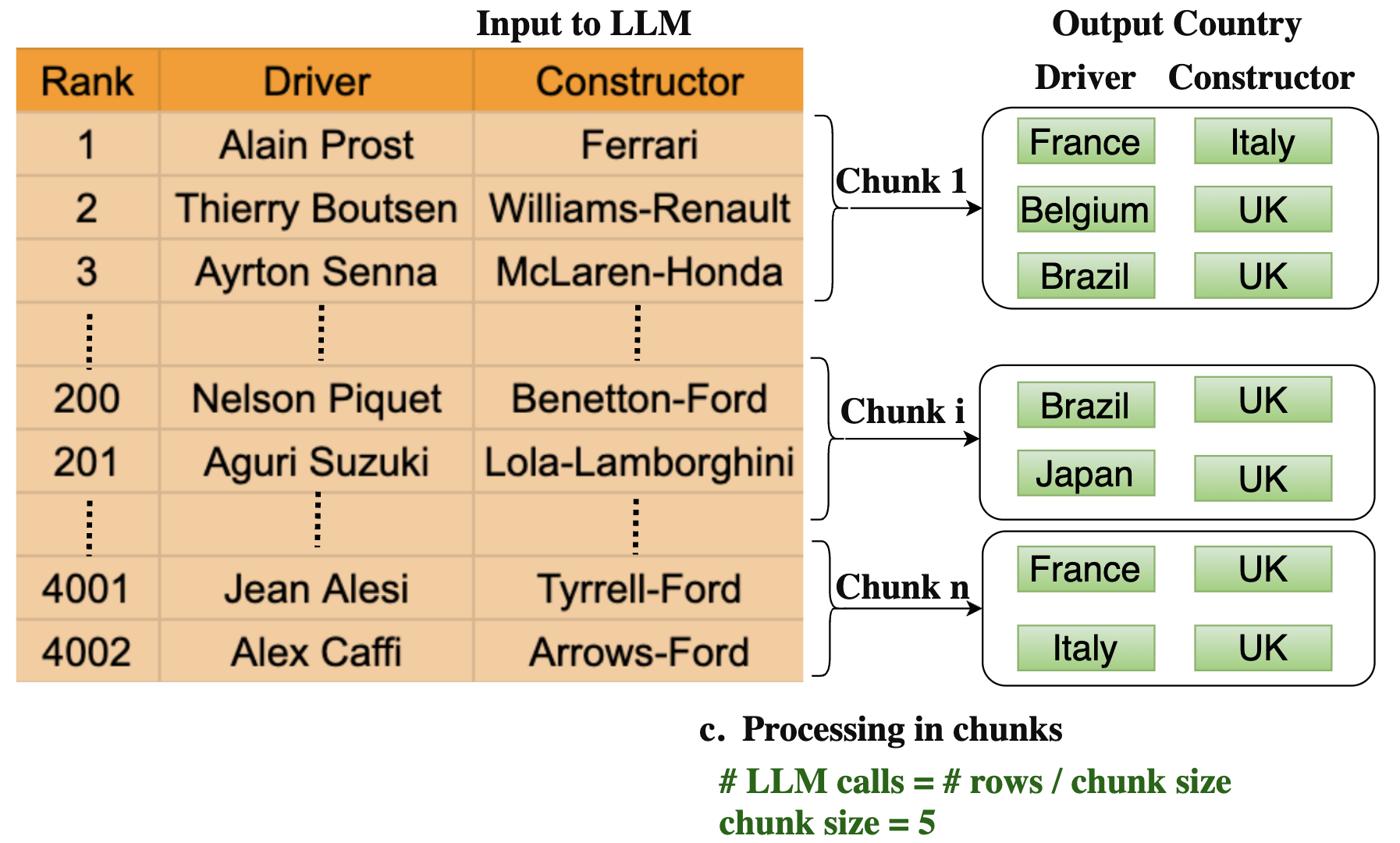}
    \caption{\small Optimizing LLM Calls: Chunk-based processing on Columns}
    \label{fig:llm-optimizaton-columns-batch}
\end{figure}

\noindent \textbf{Column-Wise Batching:}
Conventional LLM pipelines often chunk inputs row-wise, generating one column value per row across a batch. In contrast, we propose a column-wise batching strategy, depicted in Figure \ref{fig:llm-optimizaton-columns-batch}, which processes multiple columns for a small chunk of rows in a single call.
 \methodName implements this with a fixed maximum batch size defined in terms of the total number of values (columns × rows). To handle cases where the number of selected columns is large and could exceed the context length, the strategy adaptively adjusts the number of rows based on the number of columns (e.g., if more columns are included, fewer rows are processed per batch, while a smaller number of columns allows more rows). This adaptive trade-off ensures that the total input length remains manageable and consistent across diverse table structures.\\
 This approach preserves intra-row context across multiple attributes of the same entity, reducing inconsistencies that arise when attributes are generated independently. It is particularly advantageous in Retrieval-Augmented Generation (RAG) systems and memory-augmented pipelines, where repeated LLM calls over fragmented inputs can be inefficient. By extracting all relevant information in one unified query, column-wise batching lowers computational costs while maintaining high accuracy in entity-level reasoning.

\begin{lstlisting}[language=Prompt,caption={Examples of Different Plan Optimization}]
Question - The Kremlin Cup is held in Russia, and the St. Petersburg Open is also
held in Russia.

Plan:
Step 1: SQL - Filter the table to select tournaments with the names "Kremlin Cup"
and "St. Petersburg Open".

Step 2: SQL - Extract the country information from the Tournament column for the
selected tournaments.

Step 3: LLM - Summarize the results to confirm that both tournaments are held in
Russia.

Optimized Plan:
Step 1: SQL - Filter the table to select tournaments with the names "Kremlin Cup" 
and "St. Petersburg Open", and extract the country information from the Tournament 
column in a single query.

Step 2: LLM - Summarize the results to confirm that both tournaments are held in 
Russia.
\end{lstlisting}

\subsection{More Detail on handling Multimodal data}
\begin{figure}[h!]
    \centering
  \includegraphics[width=0.9\textwidth]{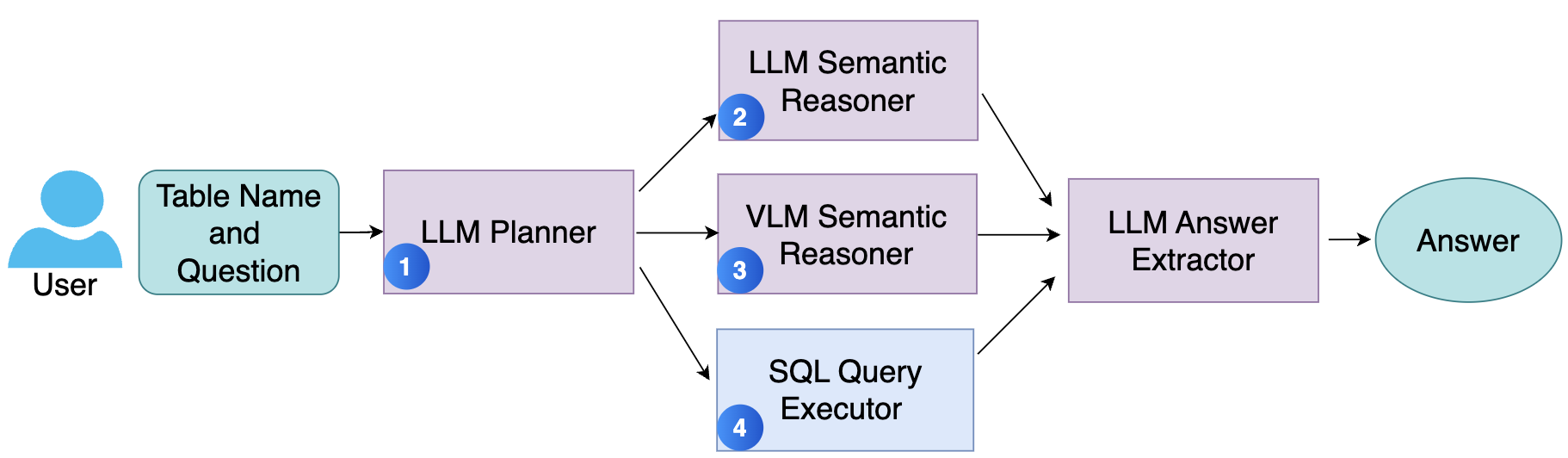}
    \caption{Modular Pipeline for query execution }
    \label{fig:opt}
\end{figure}

The proposed pipeline Figure \ref{fig:opt} is modular and designed for extensibility. Each component can be upgraded such as substituting the SQL Query Executor with an expert SQL agent to enhance execution efficiency and accuracy. Likewise, the LLM Semantic Reasoner and VLM (Vision Language Model) components can be replaced with specialized reasoning agents, allowing \methodName to adapt to evolving multimodal requirements.

\paragraph{Processing Paragraph:}

To address unstructured textual content outside the table (i.e., paragraph data), we filter these texts for relevance to both the question and the tabular data. The filtered content is used as an auxiliary knowledge source during LLM steps. This enables the system to either incorporate the external text into the tabular context or leverage it directly in the final answer generation step, depending on the Planner Agent's discretion.

\begin{figure}[h!]
    \centering
  \includegraphics[width=0.9\textwidth]{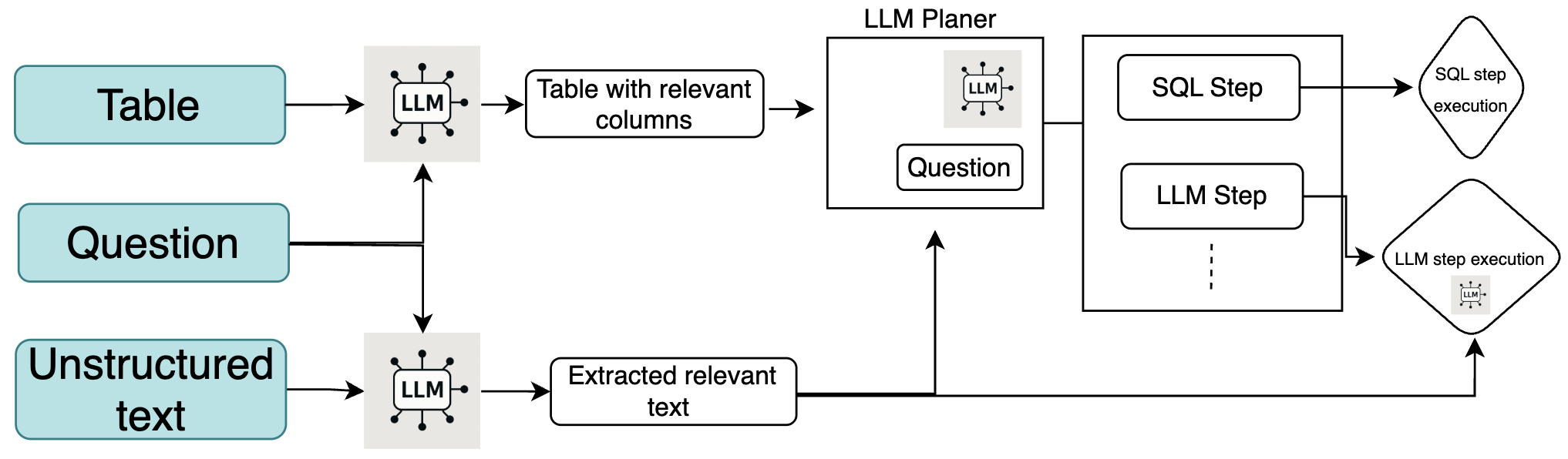}
    \caption{Paragraph-aware Weaver pipeline}
    \label{fig:paragraph}
\end{figure}

This design supports robust integration of text, table, and visual data, ensuring that the system remains scalable, accurate, and adaptable across diverse input modalities.

\subsection{Additional Analysis}
Table \ref{tab:sqlllm-steps} shows the number of LLM and SQL steps used by \methodName, across different models on WikiTQ.

\begin{table}[H]
\small
\setlength{\tabcolsep}{3pt}
\centering
\begin{tabular}{lccc}
\toprule
Steps & GPT-4o & GPT-4o-mini & Gemini-2.0-Flash   \\
\midrule
LLM Steps &46 & 178 & 122 \\
SQL Steps & 1530&1407 & 1510 \\ \midrule
Total & 1576 & 1585 & 1632\\
\bottomrule
\end{tabular}
\caption{Number of LLM steps and SQL steps used in \textit{\methodName} on WikiTQ.}
\label{tab:sqlllm-steps}
\end{table}

\paragraph{Token Usage Analysis:}
We compared the average token usage (input/output) per Table QA pair across ProTrix, H-STAR, and \methodName using three different LLM backends. As shown in Table~\ref{tab:token_usage}, \methodName maintains competitive token efficiency, especially in output size, while enabling structured multi-step reasoning. Notably, ProTrix appears lightweight due to its limited planning, whereas H-STAR and \methodName consume similar token budgets despite \methodName offering higher accuracy. Binder is excluded from this comparison due to its excessive API usage (see Table~\ref{tab:api_calls}).
\begin{table}[H]
\small
\setlength{\tabcolsep}{3pt}
	\centering
	\begin{tabular}{lccc}
		\toprule
		Token Usage/Query & GPT-4o (I/O) & GPT-4o-mini (I/O) & Gemini-2.0-Flash  (I/O) \\
		\midrule
		
		\textit{ProTrix} & 445.57 / 317.14 & 564.3 / 388.85 & 559.94 / 220.44 \\
		\textit{H-STAR} & 8,836 / 842 & 8,994 / 854 & 10,284 / 702 \\
		\textbf{\textit{\methodName}} & 8,568 / 725 & 8,723 / 796 & 6,041 / 545 \\
		
		\bottomrule
	\end{tabular}
	\caption{\small Token usage comparison per Table QA (Input/Output tokens).}
	\vspace{-1.0em}
	\label{tab:token_usage}
\end{table}

\section{Dataset Details}
\label{appendix:dataet_details}
In this section, we describe the used dataset in details. \\
\textit{- \underline{Short-Form Answering (WikiTQ)}:}
WikiTQ is a dataset designed for short-form answering where the steps to reach the answer can be relatively complex and the expected answer is a short piece of information. For example, the query “Which country had the most competitors?” in Figure \ref{fig:1}. This task is ideal for testing how well information retrieval from a semi-structured table can be handled. \\
\textit{- \underline{Fact-Checking (TabFact)}:}  
TabFact focuses on fact-checking tasks where the query involves verifying whether a particular statement is true or false. For example, the query, “Is the GDP of Japan in 2022 greater than that of Germany?” evaluates the framework’s ability to correctly interpret data points in complex tables and make valid judgments. \\
\textit{- \underline{Numerical Reasoning (FinQA)}:}
FinQA (Financial QA) is a dataset that requires the model to perform arithmetic operations or infer relationships between numerical values across different columns. For example, “What is the total revenue of Company X in 2021 after deducting expenses?” tests the model's ability to handle numerical data and apply operations such as summation, subtraction, or other mathematical reasoning tasks. \\
\textit{- \underline{Multimodal Dataset (FinQA$_{\text{MM}}$, OTT-QA$_{\text{MM}}$, MMTabQA$_{\text{MM}}$)}:}
FinQA$_{\text{MM}}$ and OTT-QA$_{\text{MM}}$ are datasets that require reasoning across both tabular data and accompanying textual context (typically a paragraph) to answer questions correctly. Unlike traditional table QA tasks, these benchmarks challenge the model to integrate information from both structured (tables) and unstructured (text) modalities. For instance, in FinQA, financial metrics are found in the table, while their definitions, dependencies, or contextual cues are only available in the surrounding text. Similarly, OTT-QA includes open-domain trivia questions, where the relevant answers often span both the table and the associated paragraphs. 

MMTabQA$_{\text{MM}}$ is a multimodal dataset that requires reasoning across both tabular data and visual information. Unlike traditional table QA tasks, it incorporates both text and images within its tables, challenging the model to integrate structured (tables), unstructured (text), and visual (images) modalities. The data set consists of query demands that combine textual descriptions, numerical data, and visual cues from images. For example, a question might ask about the relationship between financial data in the table and trends depicted in an image. MMTabQA$_{\text{MM}}$ tests the model's ability to perform complex multimodal reasoning, requiring SQL, LLM, and VLM calls to accurately derive answers. This makes it a valuable benchmark for evaluating systems that integrate and reason over multimodal inputs.

\section{Few-Shot Prompting for Plan Generation}

We use a few-shot prompting approach to generate plans during the planning stage. Through our analysis of various complex data problems, we identified three broad categories of transformation challenges that commonly arise. For each category, we manually crafted a representative example, rather than using examples from any particular dataset, to serve as in-context prompts. This was done to avoid memorization bias and to encourage cross-dataset generalization.
These examples were specifically designed to capture the reasoning skills required for hybrid queries. We will include them in the appendix for clarity and reproducibility in the final version. Below are the three categories and the corresponding examples with sample tables:

\begin{itemize}
    \item \textbf{Semantic reasoning from textual content}: These are cases where a column contains long or descriptive text, and we want the LLM to reason some semantic information which can be either direct extraction from text (e.g., extract topic from abstract text) or inference from text (e.g., infer sentiment from a text).
\end{itemize}
\begin{lstlisting}[language=Prompt,caption={Few-shot examples for semantic reasoning from textual content}]

Table: grocery_shop

item_description         sell_price            buy_price
"Indulge your senses with this botanical blend of rosemary and lavender. Gently 
cleanses while nourishing your hair, leaving it soft, shiny, and revitalized."   
7.99   4.99

Question: Which item category has the highest average profit?

Plan:
Step 1: LLM - Item_category column needs to be created using item_description column.
Step 2: SQL - Calculate average profit for each category and find the maximum.
\end{lstlisting}
\begin{itemize}
    \item \textbf{Commonsense or background knowledge inference}: In these cases, the required information is not explicitly present in the table but can be inferred using commonsense knowledge or facts the LLM is likely to have been trained on. We prompt the LLM to infer and populate a new column based on the existing data.
\end{itemize}

\begin{lstlisting}[language=Prompt,caption={Few-shot examples for commonsense or background knowledge inference}]

Table: order_delivery_history

Order ID   Product   Event              Timestamp (Local)        Location
101        Laptop    Dispatched         2025-01-14 08:00 AM      Los Angeles, USA
101        Laptop    Arrived at Hub     2025-01-15 03:00 AM      Chicago, USA

Question: Which location had the maximum time taken between dispatch at one location
and arrival or delivery at a subsequent location?

Plan:
Step 1: LLM - Convert local timestamps to UTC time for all events.
Step 2: SQL - Sort events within each Order_ID and Product by Timestamp_UTC.
Step 3: SQL - Pair Dispatched events with the corresponding Arrived at Hub or
Delivered events for each order/product and calculate the time difference.
Step 4: SQL - Display the final output with paired events, durations, and relevant
information sorted by Order_ID and Product.
\end{lstlisting}
\begin{itemize}
    \item \textbf{Pre-processing or normalization of non-SQL-friendly columns}: These are cases where a column contains values that are too diverse or unstructured to be directly used in a SQL query. Since the full table is not visible to the LLM and generating an exhaustive list of conditions is infeasible, we prompt the LLM to generate a cleaned or normalized version of the column that can be used in downstream SQL queries.
\end{itemize}

\begin{lstlisting}[language=Prompt,caption={Few-shot examples for pre-processing or normalization of non-SQL-friendly columns}]
Table: Israel at the 1972 Summer Olympics

Name                Placing
Shaul Ladani        19
Esther Shahamorov   Semifinal (5th)
Listeravov Shamil   12th

Question: Who has the highest placing rank?

Plan:
Step 1: LLM - Format the column Placing by extracting only numerical values (e.g. 5 
from Semifinal (5th)) and converting the text into numbers (e.g. Semifinal to 5, 
12th to 12, 19 to 19).
Step 2: SQL - Retrieve the highest placing (rank) from the placing column by 
selecting the minimum number in the list as lower number corresponds to higher rank.
\end{lstlisting}

\section{Additional Discussion}

\paragraph{LLM Utility in Column Transformation and Semantic Inference:}
Our core contribution lies in how we effectively leverage LLMs beyond what traditional SQL systems can offer, specifically in tasks involving column transformation and semantic inference. \methodName integrates LLM reasoning for two primary purposes: (i) handling tasks that SQL cannot perform, such as entity extraction or parsing ambiguous formats \ref{appendix:exampleTqbleQA} and (ii) inferring implicit knowledge based on pretraining, where the LLM is tasked with verifying a plan that involves semantic understanding Figure \ref{fig:main_figure}.

A detailed example from the WikiTQ dataset in Appendix \ref{appendix:exampleTqbleQA} (Listing 12) demonstrates the LLM’s capability to normalize complex date strings like \textit{Spring 1932} or \textit{1935/36} into standardized \texttt{YYYY} formats. This transformation, impossible through SQL alone, was completed through a single LLM step with high consistency. 
\begin{lstlisting}[language=Prompt,caption={Example explaination}]
Processes complex date formats (e.g., "Spring 1932", "1935/36", "1937")
Standardizes them into YYYY format (1932, 1936, 1937 respectively)
\end{lstlisting}

Additionally, our hybrid planning enables seamless coordination between SQL and LLM steps, as seen in examples involving multi-column reasoning (e.g., from a movies table with columns movie name, review content, movie information, and release date). 
\begin{lstlisting}[language=Prompt,caption={Example explaination}]
Question: "What movies suitable for kids with positive reviews should be recommended 
based on their reviews after 2018?"

Plan:
SQL: Filter movies released after 2018 using the release_date column:

SELECT * FROM movies WHERE release_date > 2018;

LLM: Utilize a Large Language Model (LLM) to evaluate whether a movie is suitable 
for children by processing:
movie_info (structured metadata)
review_content (unstructured user reviews)
new column: suitable_movies

SQL: Apply SQL filtering to retain only movies flagged as suitable by the LLM:
SELECT * FROM movies WHERE suitable_movies = 'Yes';

LLM: For the filtered movies, check which movies have positive reviews based on:
movie_info
review_content
new column: recommended

SQL: Apply SQL filtering to retain only movies which are recommended:
SELECT * FROM movies WHERE recommended = 'Yes';


\end{lstlisting}

\paragraph{Beyond Semi-structured Text:}
We define semi-structured tables following prior work \cite{infotabs}, focusing on unstructured or free-form content embedded within structured table formats, such as textual cells requiring semantic interpretation, rather than on hierarchical structures like JSON or HTML trees. While our current benchmarks focus on flat tables, the primary challenges we address stem from this embedded unstructured content. These challenges often require semantic inference, an area where traditional SQL struggles and large language models (LLMs) are critical in bridging the gap. Examples include inconsistencies in formatting, reliance on domain-specific knowledge, and implicit contextual cues necessary for accurate query interpretation.

Additionally, our approach is easily extendable to hierarchical data formats (e.g., HTML, nested JSON) through structure decoding. These formats can be transformed into a flattened, normalized table representation and then processed using \methodName. However, as we discuss later, such datasets typically lack the complex hybrid queries involving multi-step reasoning and semantic inference, which are the primary focus of our work.

\paragraph{Filtering Hybrid Queries using Binder:}
To filter hybrid queries from their original dataset counterparts, we leverage Binder’s query structure. Specifically, we identify queries that invoke any UDF, for example `QA` function as seen in the example below, which prompts the LLM with a yes/no question related to a single column. This serves as a reliable indicator that the query requires semantic reasoning beyond SQL capabilities. The example below demonstrates such a pattern, where the QA function is applied to assess contextual understanding. This filtering process ensures that the LLM is used for its intended purpose, semantic inference or interpretation, aligned with the methodology outlined in the section \ref{section:our-approach}.

\begin{lstlisting}[language=Prompt,caption={Example Binder-style UDF SQL query using QA function}]
Question: what number of games were lost at home?

UDF SQL query : 
SELECT COUNT(*) FROM w WHERE QA("map@is it a loss?"; `result/score`) = 'yes' 
AND QA("map@is it the home court of New Orleans Saints?"; `game site`) = 'yes'

\end{lstlisting}

\section{Comparison with Existing Datasets}
\label{appendix:comparison_with_dataset}

Several benchmark datasets, such as Spider \cite{spider}, HiTab \cite{hitab}, and BIRD \cite{bird} have contributed significantly to multi-hop and multi-table question answering. However, these benchmarks primarily emphasize symbolic SQL reasoning and do not align with the hybrid reasoning focus of \methodName, which requires coordinated symbolic (SQL-based) and semantic (LLM-based) reasoning.

\paragraph{Spider} was designed for cross-domain Text-to-SQL parsing over a variety of databases. While it features compositional queries, the questions can typically be answered using SQL alone. Despite recent augmentations that add paraphrasing and perturbation, the core tasks remain fully executable without any semantic interpretation, making Spider insufficient for evaluating hybrid symbolic-semantic pipelines.

\paragraph{HiTab} focuses on hierarchical tables and structural decoding challenges. Although it includes tables that require some normalization or flattening, once transformed, the resulting queries involve only 2–3 simple SQL operations (e.g., filtering, aggregation). Importantly, these tasks do not demand semantic reasoning or LLM-based operations, limiting HiTab’s suitability for hybrid evaluation.

\paragraph{BIRD} is a large-scale, multi-table QA benchmark that emphasizes compositional and multi-hop reasoning over relational tables. It serves as a rigorous testbed for symbolic systems. However, as we detail below, it lacks tasks that truly require hybrid semantic-symbolic coordination.

\subsection{Evaluation on BIRD Dataset}

We evaluate \methodName on the BIRD benchmark using GPT-4o as the backend. \methodName achieves an accuracy of 91\%, on a manually curated subset of 30 queries from the BIRD development set. We found that these sampled queries could be solved entirely  using symbolic SQL execution alone, without invoking any LLM-based semantic reasoning. These results demonstrate that while \methodName is highly effective on BIRD, the dataset’s symbolic nature limits its value for testing hybrid reasoning capabilities. Most BIRD tasks can be completed in 3–4 SQL steps involving standard operations like joins, filtering, and sorting. For instance, consider the query:

\begin{lstlisting}[language=Prompt,caption={Example execution plan}]
Question: Please list the lowest ten eligible free rates for students aged 5-17 
in continuation schools. Eligible free rates Free Meal Count (Ages 5-17) / 
Enrollment (Ages 5-17)

Plan:

SQL: Join the frpm table with the schools table using School Type and SOCType,
filtering for "Continuation High Schools".
SQL: Compute eligible free rates per school using the formula: Free Meal Count 
(Ages 5-17) / Enrollment (Ages 5-17). 
SQL: Sort schools by the calculated rate in ascending order.
SQL: Select the ten schools with the lowest eligible free rates.

SQL Queries Generated -

Step 1: SELECT f.* FROM frpm f JOIN schools s ON f.School Type = s.SOCType WHERE 
s.SOCType = 'Continuation High Schools';

Step 2: CREATE TABLE eligible_free_rates AS SELECT School Name, School Type, Free 
Meal Count (Ages 5-17), Enrollment (Ages 5-17), (Free Meal Count (Ages 5-17) / 
Enrollment (Ages 5-17)) AS eligible_free_rate FROM continuation_schools;

Step3: CREATE TABLE sorted_eligible_free_rates AS SELECT * FROM eligible_free_rates 
ORDER BY eligible_free_rate ASC;

Step 4: CREATE TABLE lowest_ten_eligible_free_rates AS SELECT * FROM
sorted_eligible_free_rates LIMIT 10;

\end{lstlisting}

Despite the multi-hop joins, these tasks do not require semantic disambiguation, commonsense reasoning, or LLM-assisted table understanding. As such, BIRD was excluded from our main hybrid benchmark comparison table, though its results highlight \methodName’s strong symbolic reasoning capabilities.

\paragraph{Why Existing Datasets Fall Short?}

While Spider, HiTab, and BIRD advance multi-hop QA in important ways, none capture the core challenges of hybrid queries where symbolic (SQL) and semantic (LLM) steps must be interleaved. \methodName is explicitly designed for such hybrid workflows, requiring intelligent planning, decomposition, and alternating execution paths capabilities not required in these prior datasets.

\subsection{Execution Strategy in the \methodName Pipeline}

\paragraph{Why SQL Execution Alone Was Insufficient?}

In \textit{\methodName}, SQL execution alone does not yield the final answer because the retrieved table may contain extraneous data or errors, particularly if any SQL-LLM steps fail or produce incomplete results. Although SQL retrieves the data, it cannot handle formatting issues, missing values, or the semantic reasoning required to extract the correct answer. Therefore, an LLM is used post-execution to refine the table, correct formatting, and extract only the relevant values, ensuring accurate final answers that SQL alone cannot guarantee.

\paragraph{Why SQL and LLM Outputs Are Not Combined?}

Unlike methods like H-STAR, which execute SQL and LLM in parallel, \textit{\methodName} selectively chooses whether SQL or LLM should handle each query part. This planning-based approach improves efficiency by avoiding the computational cost of parallel execution. As shown in Table 5: Number of API Calls Comparison per Table QA, H-STAR requires 8 LLM API calls per query, whereas \textit{\methodName} averages only 5.4, significantly reducing overhead and improving overall performance.

\subsection{Rationale for Using SQL over Python}

While Python offers greater expressiveness, we selected SQL for three key reasons that align with the objectives of our framework:

\begin{itemize}
\item \textbf{Portability:} SQL is the standard query language supported by most database engines, ensuring our approach can be easily integrated into real-world systems.
\item \textbf{Interpretability:} SQL’s declarative nature makes queries more transparent, facilitating step-by-step explainability in the planning pipeline.
\item \textbf{Efficiency:} SQL operations, such as filtering, aggregation, and joins, are highly optimized for symbolic inference over tabular data, enabling faster execution on large datasets without requiring data export.
\end{itemize}

SQL is well-suited for our use case because \textit{\methodName} is designed for seamless integration with database systems and big data environments. In contrast, Python-based solutions like Pandas can encounter scalability issues, while SQL is inherently optimized to efficiently handle large-scale tabular data.

\section{Comparison with Related Table Reasoning Methods}
\label{appeddix:comparison_with_method}

Chain-of-Table \cite{wang2024chain} represents an approach to table reasoning using iterative, chain-of-thought style planning. \methodName differs fundamentally in its methodology: it first generates a complete, high-level plan based on the question, table, and metadata, and only then begins executing each step. This global planning phase ensures interpretability and consistency across steps, similar to thinking several moves ahead in chess, rather than incrementally generating one step at a time. Because Chain-of-Table and ReAcTable both follow an incremental, step-by-step execution pattern, we included ReAcTable as a baseline. While ReAcTable captures a similar iterative reasoning paradigm, \methodName’s novelty lies in decoupling reasoning from execution through its upfront, verifiable planning.\\
The TAG framework requires manually crafted code or logic for each user query and does not support direct natural language query execution without human intervention. Comparing such methods to natural-language-driven systems like \methodName would introduce unfair bias. TAG remains a relevant and valuable framework and will serve as a future benchmark once \methodName is extended to handle multi-table logic.

\section{Future Work}
The modular design of \methodName naturally supports extensions to more complex settings. While our experiments focus on single-table English datasets to isolate hybrid reasoning capabilities, the model is not inherently limited to such settings. Preliminary tests on more complex datasets support this generalization: on BIRD, a multi-table benchmark, primarily solving tasks via only SQL, and on HiTab, which contains hierarchical tables, flattening and normalizing the schema enabled similar SQL-based execution. These results demonstrate that \methodName can adapt to multi-table and hierarchical structures when semantic inference is not required (see Appendix \ref{appendix:comparison_with_dataset}).\\
Looking forward, \methodName can be extended along three directions. First, hierarchical or nested tables could be supported by preserving context across nested attributes and extending the LLM+SQL planning strategy. Second, multi-table datasets can be handled through schema-aware planning and cross-table reasoning while maintaining hybrid execution. Third, non-English tables can be incorporated by leveraging multilingual LLMs or language-specific prompts. These directions highlight \methodName’s potential to scale seamlessly to diverse, real-world, and multilingual datasets, extending its applicability well beyond the single English tables evaluated here.

\end{document}